\documentclass{article}

    \PassOptionsToPackage{numbers, compress}{natbib}

\usepackage[preprint]{neurips_2025}

\usepackage{etoolbox}  
\newif\ifpreprint
\preprinttrue

\usepackage[utf8]{inputenc} 
\usepackage[T1]{fontenc}    
\usepackage{hyperref}       
\usepackage{url}            
\usepackage{booktabs}       
\usepackage{amsfonts}       
\usepackage{nicefrac}       
\usepackage{microtype}      
\usepackage{xcolor}         
\usepackage{amsmath}        
\usepackage{graphicx}       

\title{CCLSTM: Coupled Convolutional Long-Short Term Memory Network for Occupancy Flow Forecasting}

\author{%
  Peter Lengyel \\ 
  aiMotive \\
  Budapest, Hungary \\
  \href{https://aimotive.com}{https://aimotive.com}\thanks{Code will be available at: \href{https://github.com/aimotive/CCLSTM}{https://github.com/aimotive/CCLSTM}} \\
}

\begin{document}

\maketitle

\begin{abstract}
Predicting future states of dynamic agents is a fundamental task in autonomous driving. An expressive representation for this purpose is Occupancy Flow Fields, which provide a scalable and unified format for modeling motion, spatial extent, and multi-modal future distributions. While recent methods have achieved strong results using this representation, they often depend on high-quality vectorized inputs, which are unavailable or difficult to generate in practice, and the use of transformer-based architectures, which are computationally intensive and costly to deploy. To address these issues, we propose \textbf{Coupled Convolutional LSTM (CCLSTM)}, a lightweight, end-to-end trainable architecture based solely on convolutional operations. Without relying on vectorized inputs or self-attention mechanisms, CCLSTM effectively captures temporal dynamics and spatial occupancy-flow correlations using a compact recurrent convolutional structure. Despite its simplicity, CCLSTM achieves state-of-the-art performance on occupancy flow metrics and, as of this submission, ranks \(1^{\text{st}}\) in all metrics on the 2024 Waymo Occupancy and Flow Prediction Challenge leaderboard. For more information, visit the project website: \href{https://aimotive.com/occupancy-forecasting}{https://aimotive.com/occupancy-forecasting}
\end{abstract}

\section{Introduction}

Predicting the future states of dynamic agents is a fundamental challenge in autonomous driving. This task is complex due to several factors: it requires modeling intricate spatiotemporal dependencies and capturing long-range interactions; it is affected by contextual cues such as traffic rules and semantic signals; it must capture the inherent multi-modality of object behavior; and any practical system must be efficient enough for real-time deployment and operate robustly using only cost-effective sensor inputs like surround-view cameras and radar.

A well established representation for motion forecasting are Occupancy Flow Fields~\cite{mahjourian2022occupancy}. Occupancy grids naturally capture predictive uncertainty for object position and extent, while the associated reverse flow vectors provide temporal continuity and encode object motion. Together, these modalities offer an expressive and interpretable format for planning and control tasks in autonomous driving.

The majority of recent approaches to Occupancy Flow Field prediction~\cite{liu2025hybrid, liu2023multi, huang2022vectorflow, hu2022hope}, rely on transformer-based architectures and/or high-quality vectorized inputs. Transformer-based models are computationally intensive, which limits their practicality for deployment on resource-constrained, mass-produced onboard systems due to the associated costs. Vectorized representations must be inferred from noisy sensor data or extracted from HD maps, both of which pose a significant challenge in object detection or localization in real-world applications. Additionally, reliance on a fixed set of hand-crafted features constrains the model's ability to learn richer, potentially more informative representations from raw data.

To address these challenges, we introduce \textbf{Coupled Convolutional LSTM (CCLSTM)}, a lightweight and fully convolutional architecture designed for occupancy flow prediction. CCLSTM operates entirely in the latent space using convolutional LSTM modules that aggregate temporal context and autoregressively forecast future occupancy and flow. Our method avoids reliance on vectorized inputs and transformer components, and integrates seamlessly with existing bird's-eye view (BEV) encoder-decoder backbones, e.g., Simple-BEV~\cite{harley2023simple}, allowing end-to-end training. Our main contributions are as follows:

\begin{enumerate}
    \item \textbf{A fully convolutional LSTM-based architecture} for occupancy and flow prediction, designed for efficient spatiotemporal reasoning and real-time deployment.
    \item \textbf{A novel reverse-flow-weighted loss function} that mitigates systematic biases in occupancy flow forecasting by proportionally emphasizing dynamic objects.
    \item \textbf{State-of-the-art performance across all metrics} on the 2024 Waymo Occupancy Flow Challenge, using only rasterized inputs and no vectorized representations or pre-trained models.
\end{enumerate}

\section{Related Work}

Recurrent neural networks, particularly Long Short-Term Memory (LSTM) networks~\cite{hochreiter1997long}, have a long history in sequence modeling tasks. Sequence to Sequence (Coupled) LSTM architectures were introduced for machine translation~\cite{sutskever2014sequence}, and later adapted for unsupervised video learning~\cite{srivastava2015unsupervised}. ConvLSTM~\cite{shi2015convolutional} extended this to the spatiotemporal domain, capturing spatial correlations using convolutional gates. Our approach builds on ConvLSTM but modifies it for occupancy flow forecasting by deepening internal convolutional layers, operating in a latent space, and optimizing the convolutional LSTM equations for spatial data fusion.

Recent approaches to occupancy flow forecasting increasingly leverage transformers and vectorized representations to exploit global receptive fields and utilize multi-modal inputs. STrajNet~\cite{liu2023multi} combines rasterized feature maps with vectorized trajectories, employing attention mechanisms for vector encoding, spatiotemporal fusion and spatial reasoning. Concurrently, VectorFlow~\cite{huang2022vectorflow} proposes a CNN-based encoder-decoder that combines vectorized and visual features through cross-attention modules. HGNET~\cite{chen2024hgnet} adopts a transformer-based architecture for both vector and raster modalities, introducing a Feature-Guided Attention (FGAT) module for spatial fusion, and a GRU-based module for temporal prediction. DOPP, a variant of HPP~\cite{liu2025hybrid} by the same authors, employs Ms-OccFormer, a custom multi-transformer cascade decoder, to iteratively predict future marginal-conditioned occupancy. Like other recent approaches, DOPP leverages both vectorized and visual features.

CCLSTM differs from these solutions significantly, the key differences being avoiding the use of both vectorized inputs and transformer architectures. Vectorized inputs are precise but difficult to obtain reliably in real-world settings. Inferring them from sensor data introduces noise and error, while HD maps depend on accurate localization, which is not always available. Moreover, vectorized formats constrain learning to a fixed set of engineered features. Our approach bypasses these issues by training directly on rasterized BEV data, learning rich features from raw inputs and enabling deployment in more diverse and uncertain environments. While transformers are effective for temporal prediction, their computational cost grows rapidly with spatiotemporal data due to their global receptive field. Techniques like Shifted Windows (Swin) used by DOPP~\cite{liu2025hybrid} or Deformable Attention (DAT) used by STrajNet~\cite{liu2023multi} reduce this cost by optimizing the receptive field of attention, but models still remain resource-intensive and often require specialized operations not supported by the NPUs used in embedded systems.

\section{Methodology}

\subsection{Model}
\label{sec:Model}

\begin{figure}
  \centering
  \includegraphics[width=1.0\textwidth]{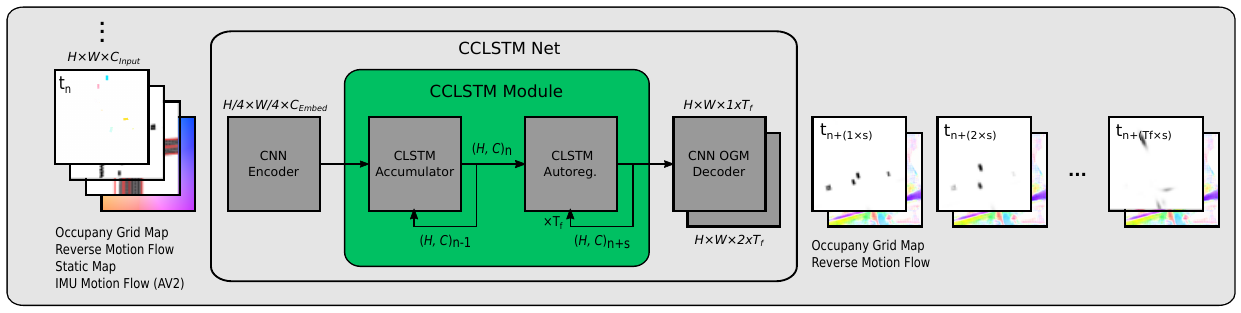}
  \caption{An overview of CCLSTM. Rasterized input grids are concatenated along the channel dimension and
  encoded via a CNN. The encoded features are aggregated via the accumulator CLSTM. The hidden and cell states
  of the accumulator CLSTM are used to initialize the forecasting CLSTM. The forecasting CLSTM is then
  autoregressively called to predict encoded futures states. The future hidden states are then passed to a
  CNN Decoder, to produce occupancy and flow grids.}
  \label{fig:architecture}
\end{figure}

Our model is a fully convolutional architecture composed exclusively of $3{\times}3$ and $5{\times}5$ convolutional layers, resulting in a compact design with just 31M learnable parameters. This efficiency stems from extensive parameter reuse across both the recurrent accumulation and autoregressive prediction stages. While convolutions have limited receptive fields, which can be suboptimal for modeling large spatial interactions, our design mitigates this by decomposing lateral feature movement into smaller, incremental steps, enabling effective spatial reasoning through iterative updates (see Fig.~\ref{fig:architecture}).

\textbf{Encoder:} The encoder consists of \( 4 \) convolutional layers where the first layer uses a kernel size of \( 5 \) and the remaining \( 3 \) layers use a kernel size of \( 3 \). All layers are configured without bias and are followed by a leaky-ReLU activation and channel-wise group normalization. This module progressively downsamples the spatial resolution by a factor of \( 4 \). The input channel dimension depends on the number of concatenated feature maps, while the output embedding dimension is empirically set to \( C = 256 \).

\textbf{Decoder:} The model uses two parallel decoder branches: one for occupancy prediction and one for reverse flow. Each branch consists of \( 3 \) transposed convolutional layers, followed by leaky ReLU activations and channel-wise group normalization, and ends with a convolutional layer for output smoothing. During inference, a sigmoid activation is applied to the occupancy decoder output. Both decoders upsample the latent features back to the original input resolution.

\textbf{CCLSTM:} The equations of the CLSTM modules are shown in Eq.~\ref{eq:clstm_accumulation} and Eq.~\ref{eq:clstm_forecast} respectively, where \( \ast \) denotes the convolution operator, \( \circ \) denotes the Hadamard product and \(\|\) denotes channel-wise concatenation. The terms \( W \) and \( b \) refer to the convolutional weights and biases, respectively. The subscripts \( i \), \( f \), \( g \) and \( o \) denote the LSTM's \( input \), \( forget \), \( gate \) and \( output \) gates. In practice, each convolutional operation is implemented using a small convolutional neural network rather than a single layer, comprised of \( 3 \) convolutional layers of kernel size \( 3 \), interleaved with leaky-relu activations. Channel-wise group normalization is used to stabilize training.

\textbf{Accumulation CLSTM Cell:} The Accumulation CLSTM Cell, defined in Eq.~\ref{eq:clstm_accumulation}. incrementally integrates incoming latent feature maps \( X_t \) previously aggregated hidden and cell states \( C_{t-1} \), \( H_{t-1} \). This process is iterated over the observed sequence with the time step $\Delta t$ typically aligned with the sensor frame rate.

\begin{equation}
\begin{aligned}
    i_t &= \sigma(W_{i} * [X_t \| H_{t-1}] + b_i); \\
    f_t &= \sigma(W_{f} * [X_t \| H_{t-1}] + b_f); \\
    g_t &= \tanh(W_{g} * [X_t \| H_{t-1}] + b_g); \\
    o_t &= \sigma(W_{o} * [X_t \| H_{t-1}] + b_o); \\
    C_t &= f_t \circ C_{t-1} + i_t \circ g_t; \\
    H_t &= o_t \circ \tanh(\mathrm{GroupNorm}(C_t))
\end{aligned}
\label{eq:clstm_accumulation}
\end{equation}

\textbf{Forecasting CLSTM Cell:} The Forecasting CLSTM cell, specified in Eq.~\ref{eq:clstm_forecast}, initializes its hidden and cell states with the final states from the accumulator module, denoted by \( C^{\text{acc}}_t \), \( H^{\text{acc}}_t \). It is then autoregressively unrolled to predict future latent states. Notably, the forecasting module may operate with a different temporal resolution $\Delta t$ than the accumulation module. The neural network implementing the weights uses convolutional layers of kernel size \( 5 \). The predicted latent features are subsequently passed through the aforementioned Decoder network to generate future occupancy and flow fields.

\begin{equation}
\begin{aligned}
    i_t &= \sigma(W_{i} * H_{t-1} + b_i)  \\
    f_t &= \sigma(W_{f} * H_{t-1} + b_f) \\
    g_t &= \tanh(W_{g} * H_{t-1} + b_g) \\
    o_t &= \sigma(W_{o} * H_{t-1} + b_o) \\
    C_t &= f_t \circ C_{t-1} + i_t \circ g_t, \quad \text{where } C_{-1} =  C^{\text{acc}}_0 \\
    H_t &= o_t \circ \tanh(\mathrm{GroupNorm}(C_t)), \quad \text{where } H_{-1} = H^{\text{acc}}_0
\end{aligned}
\label{eq:clstm_forecast}
\end{equation}

\subsection{Loss}
\label{sec:Loss}

We train our model by minimizing three loss terms: occupancy loss \( \mathcal{L}_{\text{occupancy}} \), flow loss \( \mathcal{L}_{\text{flow}} \), and trace loss \( \mathcal{L}_{\text{trace}} \). The total loss function is defined as:

\begin{equation}
\mathcal{L} = \lambda_{\text{occupancy}} \mathcal{L}_{\text{occupancy}} + \lambda_{\text{flow}} \mathcal{L}_{\text{flow}} + \lambda_{\text{trace}} \mathcal{L}_{\text{trace}}
\end{equation}

where \( \lambda_{\text{occupancy}} \), \( \lambda_{\text{flow}} \), and \( \lambda_{\text{trace}} \) are hyperparameters used to balance the contributions of each loss term. We empirically set these to \( \lambda_{\text{occupancy}} = 1000 \), \( \lambda_{\text{flow}} = 25 \), and \( \lambda_{\text{trace}} = 10 \).

Let \( O \in \{0, 1\}^{T \times 1 \times H \times W} \) be the expected occupancy, 
\( \hat{O} \in \mathbb{R}^{T \times 1 \times H \times W} \) be the predicted occupancy logits, 
\( F \in \mathbb{R}^{T \times 2 \times H \times W} \) the expected flow, 
\( \hat{F} \in \mathbb{R}^{T \times 2 \times H \times W} \) be the predicted flow.

\begin{figure}
  \centering
  \includegraphics[width=1.0\textwidth]{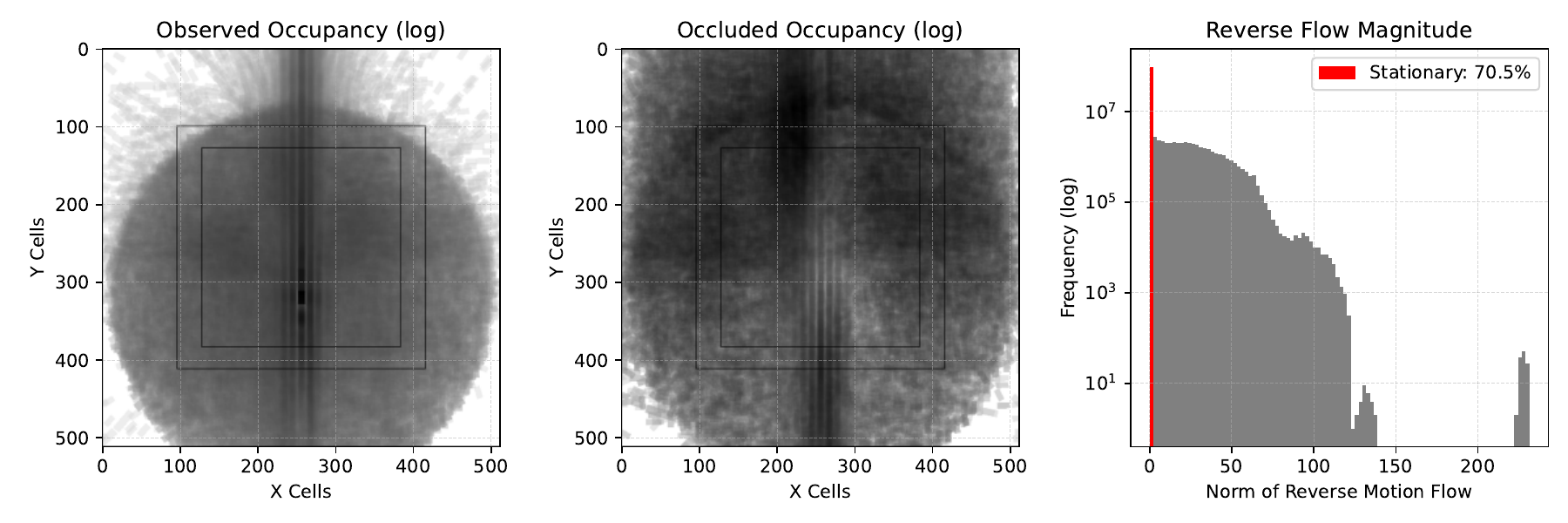}
  \caption{\textbf{Waymo Open Motion Dataset validation set distribution analysis:} Visualizing the expected (future) observed and occluded occupancy distribution indicate that data diversity can be increased by using larger FoV rasters: (W, H) = 256, 320 and 512. In the case of observed occupancy, some distribution skew may be observed due to a circular pattern, assumed to be a data collection methodology artifact. Visualizing the magnitude (norm) of the occupied cell's reverse motion flow as a histogram, shows a significant bias for stationary objects.}
  \label{fig:womd_distribution}
\end{figure}

\begin{figure}
  \centering
  \includegraphics[width=1.0\textwidth]{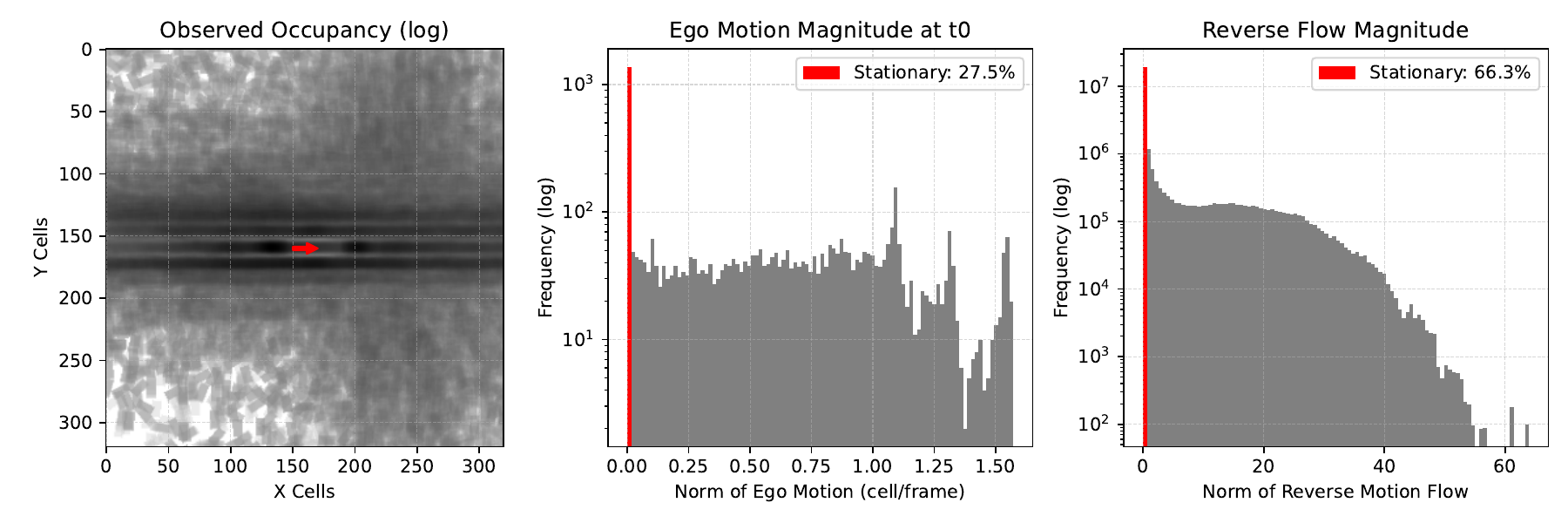}
  \caption{\textbf{Argoverse 2 validation set (initial 5{,}000 samples) analysis:} The expected future occupancy distribution resembles that of WOMD, despite using ego-centered coordinates (red arrow indicates ego-vehicle direction). At prediction time, the ego-vehicle is stationary in approximately 28\% of cases. Although the use of a ego-centered coordinate system causes stationary objects to appear in motion during aggregation, the predicted occupancy in the frame at \(t_0\) keeps truly stationary objects fixed. This explains the similarity in reverse flow magnitude distributions across datasets.}
  \label{fig:argoverse_distribution}
\end{figure}

\textbf{Occupancy Loss:} We compute the loss as a weighted sum of \texttt{BCEWithLogitsLoss} over the temporal and spatial dimensions, where \( F \) is the expected reverse motion flow and \( \alpha \in \mathbb{R} \) a scaling factor empirically set to 10. Using the norm of \( F \) for weighting is intended to compensate for the observation that the dataset is significantly biased toward stationary objects (see Fig.~\ref{fig:womd_distribution}, Fig.~\ref{fig:argoverse_distribution}).

\begin{equation}
\mathcal{W}_{t,h,w} = O_{t,h,w} \cdot \left( \frac{\|F_{t,h,w}\|}{\alpha} + 1.0 \right)
\end{equation}

\begin{equation}
\mathcal{L}_{\text{occupancy}} = \frac{1}{thw} \cdot \sum_{t,h,w} (\texttt{BCEWithLogitsLoss}(\hat{O}_{t,h,w}, O_{t,h,w}) \cdot (\mathcal{W}_{t,h,w} + 1.0))
\end{equation}

\textbf{Flow Loss:} We use observed occupancy weighted MSE loss for predicted motion flow.

\begin{equation}
\mathcal{\alpha} = \sum_{t,h,w} \mathcal{O}_{t,h,w}^{obs}
\end{equation}

\begin{equation}
\mathcal{L}_{\text{flow}} = \frac{1}{\alpha} \cdot \sum_{t,h,w} \mathcal{O}_{t,h,w}^{obs} \cdot \texttt{L1Loss}(\mathcal{\hat{F}}_{t,h,w}, \mathbb{F}_{t,h,w})
\end{equation}

\textbf{Traced Loss:} We reinforce the Flow loss with a Trace loss to strengthen consistency using adjacent occupancy ground truth grids \( \mathcal{O}^{k-1} \) and \( \mathcal{O}^{k} \), where \( \circ \) denotes the function application (warping) of \( \mathcal{\hat{F}}^{k} \) to transform \( \mathcal{O}^{k-1} \)

\begin{equation}
\mathcal{\alpha} = \sum_{t,h,w} \mathcal{O}_{t,h,w}^{k}
\end{equation}

\begin{equation}
\mathcal{L}_{\text{trace}} = \frac{1}{\alpha} \cdot \sum_{t,h,w} \texttt{MSELoss}(\mathcal{O}_{t,h,w}^{k} \cdot (\mathcal{\hat{F}}_{t,h,w}^{k} \circ \mathcal{O}_{t,h,w}^{k-1}), \mathcal{O}_{t,h,w}^{k})
\end{equation}

\subsection{Dataset}
\label{sec:Dataset}

\subsubsection{WOMD (Occupancy and Flow Prediction)}

The Waymo Open Motion Dataset (WOMD)~\cite{ettinger2021large} comprises 485{,}568 training, 4{,}400 validation, and 4{,}400 test samples. Historical agent states are sampled at 10\,Hz over the past 1 second (\(T_h = 10\)), and the forecasting objective is to predict future occupancy and flow over the next 8 seconds at 1\,Hz (\(T_f = 8\)). Both input and output rasters have a resolution of \(H, W = 320\), corresponding to a \(100 \times 100\)\,m\(^2\) area in real-world coordinates. For our challenge submission we use \(H, W = 512\), corresponding to a \(160 \times 160\)\,m\(^2\) area in real-world coordinates. A central crop of \(H, W = 256\) is used for evaluation.

\textbf{Input (\( \boldsymbol{X} \)):} Following STrajNet~\cite{liu2023multi} and OFMPNet~\cite{murhij2024ofmpnet}, the input rasters consist of historical and current: (1) occupancy grids \( O_t \in \mathbb{R}^{H \times W \times 1} \); (2) dense semantic maps \( M_t \in \mathbb{R}^{H \times W \times 3} \), which encode road topology and traffic light states as RGB images; and (3) backward flow fields \( F_t \in \mathbb{R}^{H \times W \times 2} \), derived from agent displacements between successive occupancy frames. The input sequence spans timesteps \( t \in [-T_h + 1, 0] \), where the initial frame at \( t = -T_h \) is excluded due to the flow computation requiring a preceding frame. Note that the dataset is rendered in a static frame of reference, and thus the semantic map \( M \) remains constant over time.

\textbf{Expected Output ($\boldsymbol{Y}$):} Inline with preceding solutions, the ground-truth rasters consist of future: (1) observed occupancy \( O_t \in \mathbb{R}^{H \times W \times 1} \); (2) occluded occupancy \( O_t \in \mathbb{R}^{H \times W \times 1} \); and (3) backward flow fields \( F_t \in \mathbb{R}^{H \times W \times 2} \). The Output sequence spans timesteps \( t \in [1, T_f] \).

\subsubsection{AV2 (Motion Forecasting Dataset)}

Argoverse 2 Motion Forecasting Dataset (AV2)~\cite{wilson2023argoverse} comprises 200{,}000 training, 25{,}000 validation, and 25{,}000 test samples. Historical agent states are sampled at 10\,Hz over the past 5 seconds (\(T_h = 50\)), and the forecasting objective is to predict future occupancy and flow over the next 6 seconds at 1.667\,Hz (\(T_f = 10\)). Both input and output rasters have a resolution of \(H, W = 320\), corresponding to a \(80 \times 80\)\,m\(^2\) area in real-world coordinates. We exclude the rendering of the track-id  \texttt{"AV"} (ego-vehicle), all non \texttt{VEHICLE} object-types and the \texttt{TRACK\_FRAGMENT} track-category.

\textbf{Input ($\boldsymbol{X}$):} The input rasters consist of historical and current (1) occupancy grids \( O_t \in \mathbb{R}^{H \times W \times 1} \); (2) lane occupancy grids \( L_t \in \mathbb{R}^{H \times W \times 1} \); and (3) rasterized egomotion flow \( E_t \in \mathbb{R}^{H \times W \times 2} \), computed from ego displacements across frames. The input sequence spans timesteps \( t \in [-T_h + 1, 0] \), where the initial frame at \( t = -T_h \) is excluded due to the flow computation requiring a preceding frame. Rasterized egomotion is defined as the reverse flow for all grid elements \( E_t \in \mathbb{R}^{H \times W \times 2} \) between consecutive frames. Note that the dataset is rendered in a ego-centric frame of reference, and thus the semantic map \( L \) is not necessarily constant across time.

\textbf{Expected Output ($\boldsymbol{Y}$):} The ground-truth rasters consist of future: (1) occupancy \( O_t \in \mathbb{R}^{H \times W \times 1} \); and (2) backward flow fields \( F_t \in \mathbb{R}^{H \times W \times 2} \). The output sequence spans timesteps \( t \in [1, T_f] \). The outputs \( t \in [1, T_f] \) are rendered in the ego-centric frame of reference at \( t = 0 \). For consistency with WOMD, we use the initial 5{,}000 samples of the sorted validation set for evaluation.

\subsection{Training}
\label{sec:Training}

CCLSTM is trained end-to-end, from scratch for 10 epochs with a batch size of 32 on a single NVIDIA A100 GPU. Optimization is performed using AdamW with an initial learning rate of 0.002. A cosine annealing scheduler is employed, configured with \( T_{\text{mult}} = 1 \) and \( \eta_{\text{min}} = \text{lr} / 100 \). The hidden states of the Accumulation CLSTM at \( t = T_h + 1 \) are initialized with 0. We use backpropagation through time, and performance-based checkpointing.

\textbf{WOMD:} To reduce overfitting and enhance data diversity, we apply random 180° rotations and train using an increased raster size of \( H, W = 320 \). The output is cropped to the Occupancy Flow Fields Challenge submission's expected RoI of \( H, W = 256 \). This approach is consistent with prior works that leveraged vectorized inputs beyond the raster RoI, and is justified by the RoI provided by surround-view sensor inputs (e.g., cameras) in practical deployments. During training, we accumulate the input sequence over \( t \in [-T_h + 1, 0] \), and perform forecasting only for the final timestep \( t = 0 \). The forecasted frames comprise \( t \in [1, T_f] \). For our challenge submission, we use a further increased raster size of \( H, W = 512 \), a batch size of 8 and scale the initial learning rate to 0.00005. No pre-training, model ensembling, test-time augmentation or external data beyond WOMD is used.

\textbf{AV2:} Due to memory constraints and consistency with WOMD, we accumulate the input sequence over \( t \in [-10, 0]\), and perform forecasting only for the final timestep \( t = 0 \). The predicted frames comprise \( t \in [1, T_f] \). We augment the data by uniformly sampling \( t \in [T_h, T_f] \) from the full sequence length of \( t \in [0, 110] \). No pre-training, model ensembling, test-time augmentation or external data beyond Argoverse 2 is used.

\section{Experiments}

We evaluate our method on the official Waymo Occupancy Flow Challenge test set for comparison with existing approaches using the standard metrics proposed in the challenge~\cite{mahjourian2022occupancy}. Ablation studies are conducted on the corresponding validation set to investigate key design choices. Additionally, we report results on an ego-centric rasterization of the Argoverse 2 dataset.

\begin{figure}
    \label{fig:qualitative_results}
    \centering
    \begin{minipage}{0.49\textwidth}
        \centering
        \includegraphics[width=\textwidth]{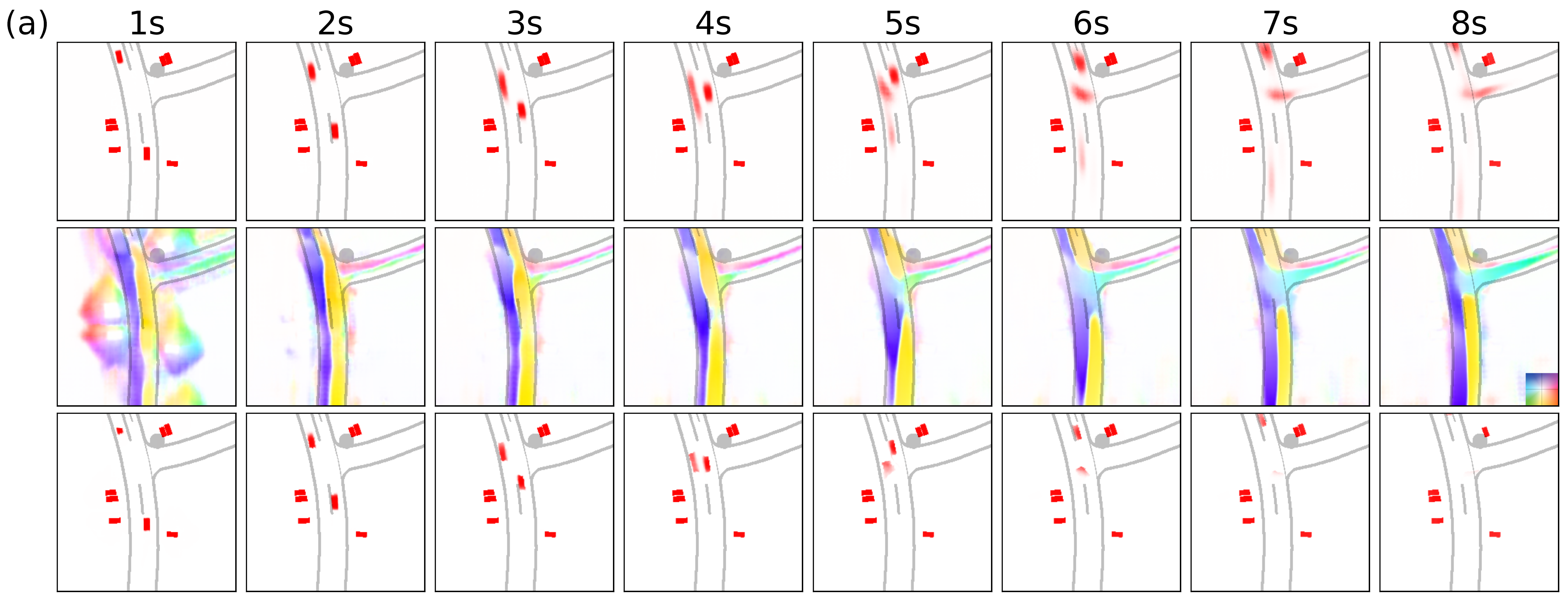}
    \end{minipage} 
    \begin{minipage}{0.49\textwidth}
        \centering
        \includegraphics[width=\textwidth]{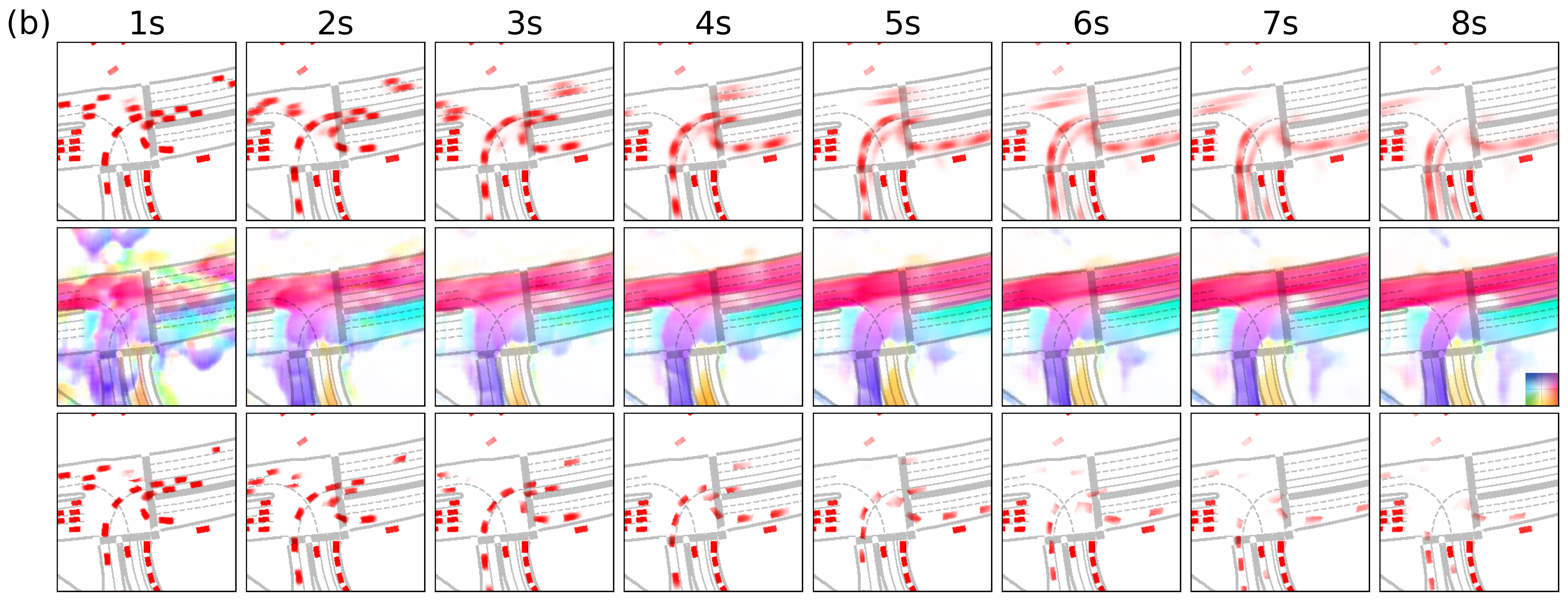}
    \end{minipage}
    \begin{minipage}{0.49\textwidth}
        \centering
        \includegraphics[width=\textwidth]{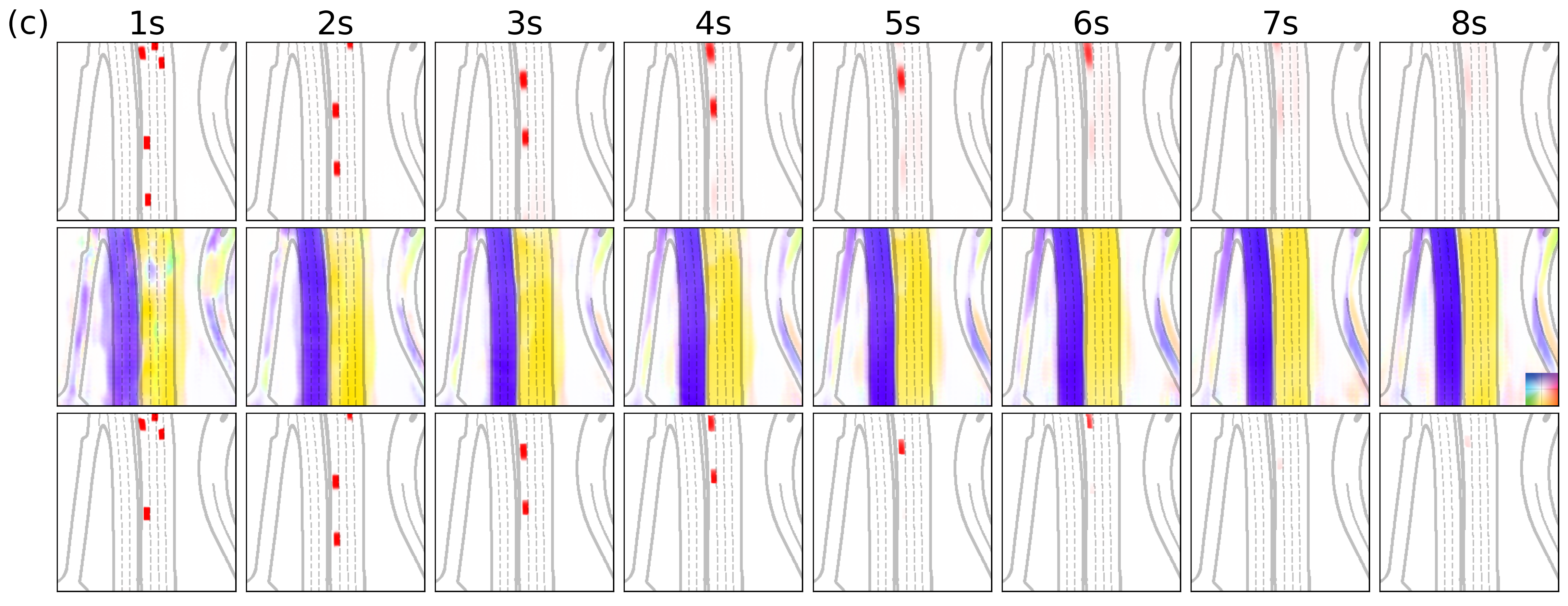}
    \end{minipage}
    \begin{minipage}{0.49\textwidth}
        \centering
        \includegraphics[width=\textwidth]{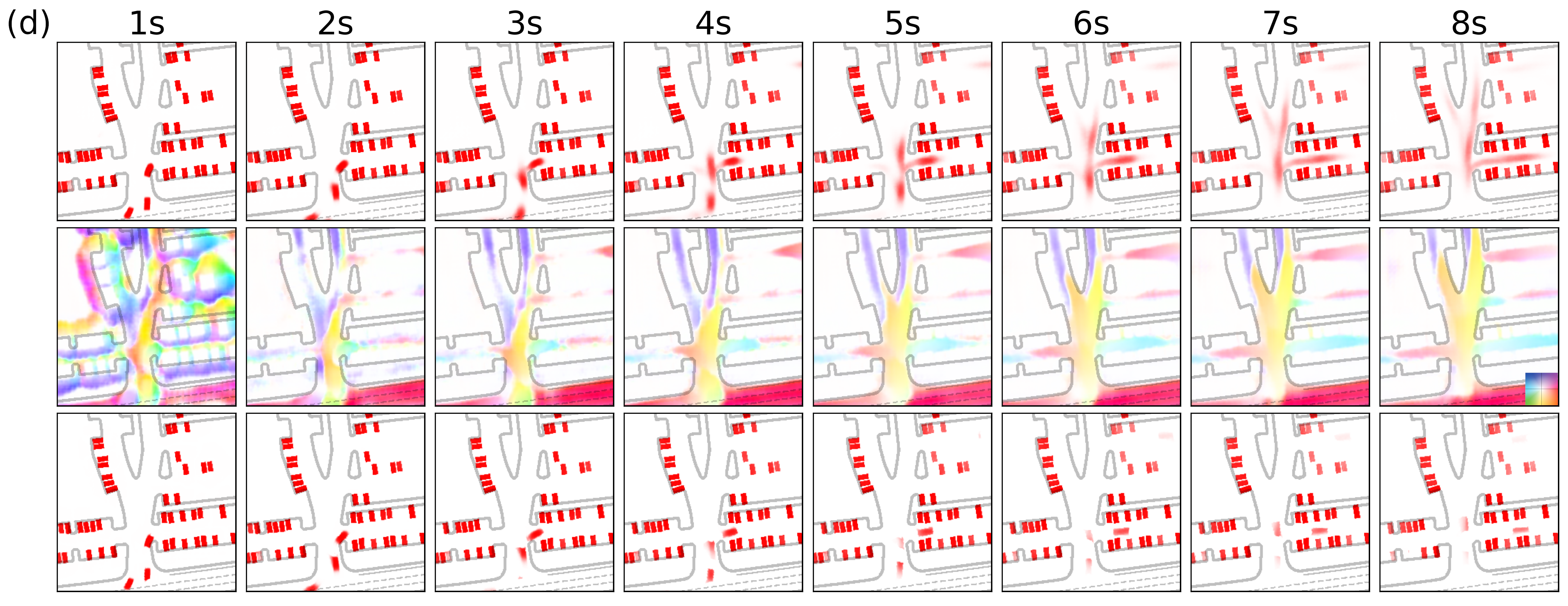}
    \end{minipage}
    \caption{\textbf{Qualitative results on WOMD validation set}. Each subplot displays the testing result of (1) occupancy, (2) backward flow, and (3) flow-traced occupancy. Scenarios: (a) Multi-model agent Interaction; (b) U-Turn; (c) Fast moving agent; (d) Agent separation in dense traffic.}
\end{figure}

\subsection{Waymo Open Motion Dataset (WOMD) Results}
\label{sec:womd_results}

We evaluate our method on the Waymo Open Motion Dataset Occupancy and Flow Prediction Challenge test set via the online evaluation server (Tab.~\ref{tab:test_benchmarks}). We compare against other state-of-the-art models, including DOPP~\cite{liu2025hybrid}, STrajNet~\cite{liu2023multi}, and VectorFlow~\cite{huang2022vectorflow}, while excluding methods that leverage pre-trained encoders, such as HOPE~\cite{hu2022hope}. Unlike these methods, however, our approach does not utilize vectorized inputs, relying solely on rasterized feature maps, yet still achieves comparable or superior performance.

\begin{figure}
  \centering
  \includegraphics[width=1.0\textwidth]{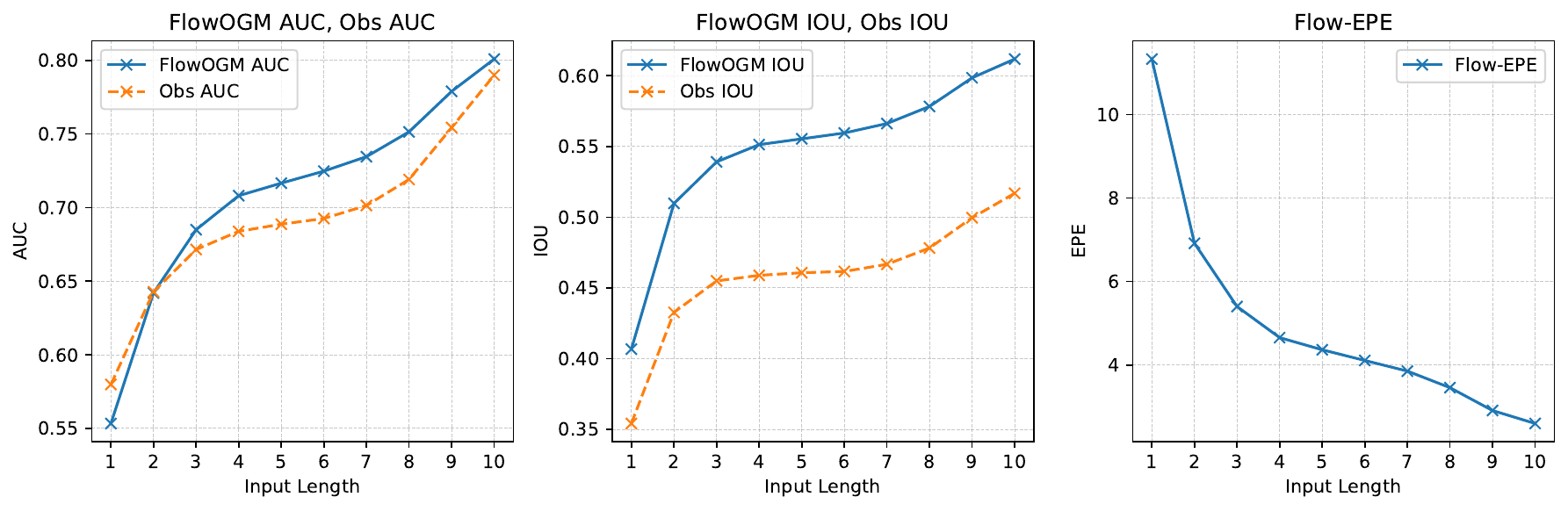}
  \caption{\textbf{WOMD inference length analysis:} Validation metrics plotted as a function of input sequence length. The results demonstrate that longer input sequences lead to improved predictive accuracy, emphasizing the recurrent module’s capacity for temporal data fusion}
  \label{fig:sequence_waymo}
\end{figure}

\begin{figure}
  \centering
  \includegraphics[width=1.0\textwidth]{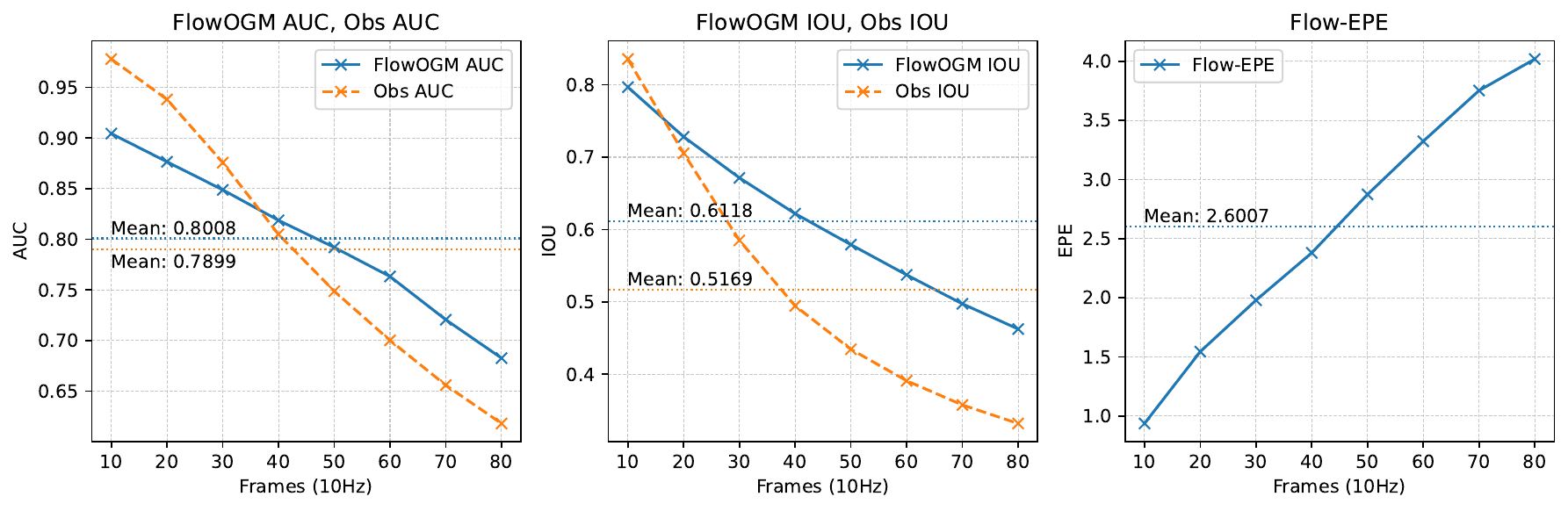}
  \caption{\textbf{WOMD metrics per-waypoints results:} Validation metrics as a function of forecast horizon. The plot visualized the degradation of metrics over longer forecast horizons due to increased uncertainty.}
  \label{fig:curve_plot_waymo}
\end{figure}

Unlike other methods, which accumulate a limited sliding window of past data, our solution is designed
to accumulate data from an arbitrary length sequence without an increase in computation,
which is a realistic use case in AVs. To analyze the relationship between inference sequence length and performance, we evaluate our model on a range of input frames (from 1 to 10), which is the maximum permitted by the WOMD dataset (Fig.~\ref{fig:sequence_waymo}). We also provide metrics per predicted timestep in a curve plot (Fig.~\ref{fig:curve_plot_waymo}), as this is vital information for evaluating usability. Qualitative results showcasing complex agent interaction modeling and multi-modal future prediction are available in Fig.~\ref{fig:qualitative_results} and Appendix~\ref{appendix:additional_qualitative_results}.

\begin{table}
  \caption{\textbf{WOMD:} Comparison of Different Models on Occupancy and Flow Prediction (test set)}
  \label{tab:test_benchmarks}
  \centering
  \small
  \begin{tabular}{lccccccc}
    \toprule
    {Model} & \multicolumn{2}{c}{Observed} & \multicolumn{2}{c}{Occluded} & & \multicolumn{2}{c}{Flow-Grounded} \\
    \cmidrule(r){2-3} \cmidrule(r){4-5} \cmidrule(r){7-8}
    & AUC $\uparrow$ & Soft IoU $\uparrow$ & AUC $\uparrow$ & Soft IoU $\uparrow$ & Flow EPE $\downarrow$ & AUC $\uparrow$ & Soft IoU $\uparrow$ \\
    \midrule
    DOPP            & \textit{0.7972} & 0.3429 & \textit{0.1937} & 0.0241 & \textit{2.9574} & \textit{0.8026} & 0.5156 \\
    STrajNet        & 0.7514 & 0.4818 & 0.1610 & 0.0183 & 3.5867 & 0.7772 & \textit{0.5551} \\
    VectorFlow      & 0.7548 & \textit{0.4884} & 0.1736 & \textit{0.0448} & 3.5827 & 0.7669 & 0.5298 \\
    STNet           & 0.7552 & 0.2299 & 0.1658 & 0.0180 & 3.3779 & 0.7564 & 0.4431 \\
    HGNET           & 0.7332 & 0.4211 & 0.1656 & 0.0389 & 3.6699 & 0.7403 & 0.4498 \\
    CCLSTM (Ours)   & \textbf{0.8154} & \textbf{0.5321} & \textbf{0.2077} & \textbf{0.0606} & \textbf{2.6831} & \textbf{0.8196} & \textbf{0.6256} \\
    \bottomrule
  \end{tabular}
\end{table}

\subsection{Argoverse 2 (AV2) Results}

\begin{table}
  \caption{\textbf{AV2:} Ablation Study for input choices (validation set)}
  \label{tab:argoverse_ablation}
  \centering
  \small
  \begin{tabular}{lccccccc}
    \toprule
    {Model} & \multicolumn{2}{c}{Observed} & \multicolumn{2}{c}{Occluded} & & \multicolumn{2}{c}{Flow-Grounded} \\
    \cmidrule(r){2-3} \cmidrule(r){4-5} \cmidrule(r){7-8}
    & AUC  & Soft IoU  & AUC  & Soft IoU & Flow EPE & AUC & Soft IoU  \\
    \midrule
    CCLSTM & 0.7143 & 0.3937 & 0.0 & 0.0 & 1.2390 & 0.8210 & 0.5996 \\
    CCLSTM-IMU & 0.7154 & 0.3960 & 0.0 & 0.0 & 1.2280 & 0.8209 & 0.5943 \\
    \bottomrule
  \end{tabular}
\end{table}

\begin{figure}
  \centering
  \includegraphics[width=1.0\textwidth]{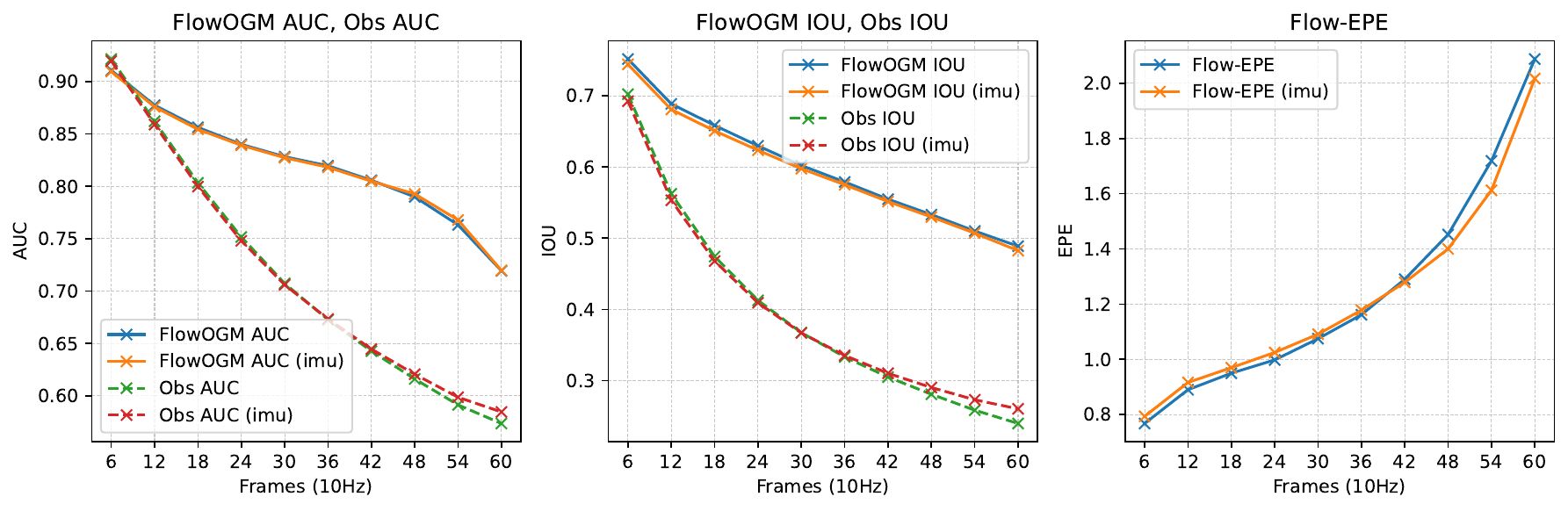}
  \caption{\textbf{AV2 Dataset metrics per-waypoints results:} Validation metrics as a function of forecast horizon. The plot visualized the degradation of metrics over longer forecast horizons due to increased uncertainty. The ablation using rasterized IMU data demonstrates improved performance, with the effect becoming more pronounced at longer forecast horizons.}
  \label{fig:curve_plot_av2}
\end{figure}

\begin{figure}
  \centering
  \includegraphics[width=1.0\textwidth]{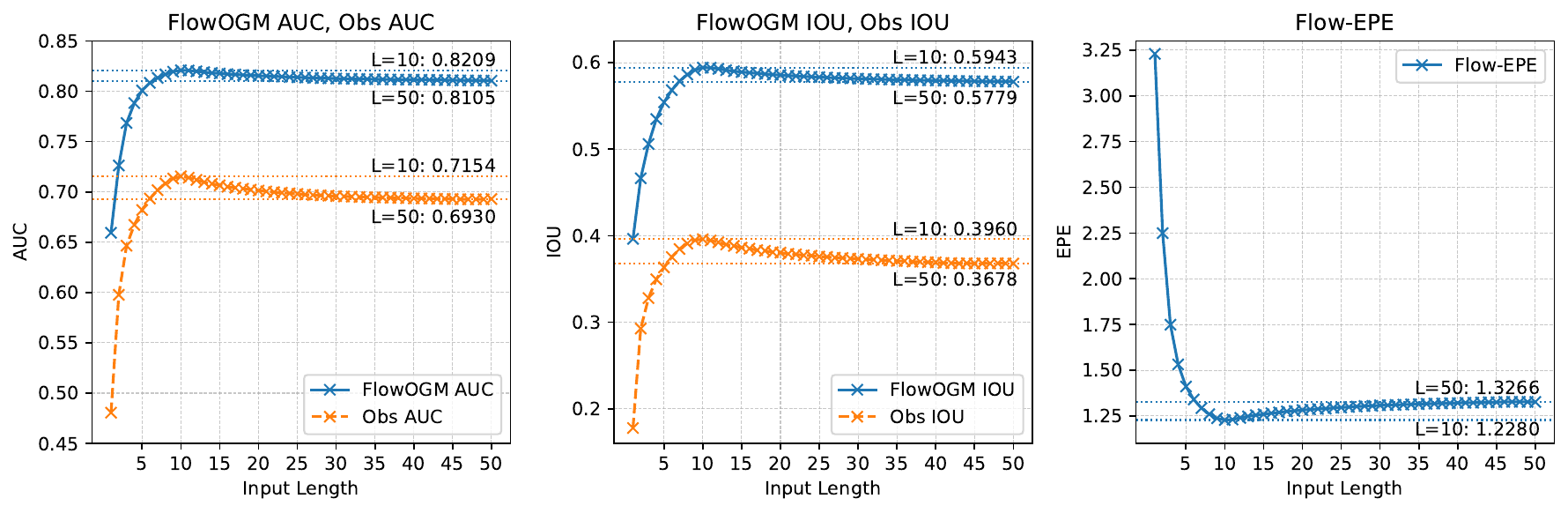}
  \caption{\textbf{AV2 inference length analysis:} Validation metrics plotted as a function of input sequence length show a drop in performance for sequence lengths longer than the training sequence length (10 frames).}
  \label{fig:sequence_av2}
\end{figure}

We report results of our method on the ego-centric rasterization of Argoverse 2. We use the same metrics and framework as for the evaluation of WOMD. While it is not possible to directly compare results between the WOMD stationary and the AV2 ego-centric frame of reference dataset, the results in Tab.~\ref{tab:argoverse_ablation} and Fig.~\ref{fig:curve_plot_av2} indicate that the method generalizes to a moving coordinate system. To examine the relationship between sequence length and performance, we evaluate our model on a range of input sequence lengths (from 1 to 50); see Fig.~\ref{fig:sequence_av2}.. The trend of this data shows that the model performance slightly degrades for input sequence lengths longer than the training sequence length.

\subsection{Ablation Study}

\textbf{WOMD:} We conduct an ablation study to quantify the contribution of key architectural components Tab.~\ref{tab:womd_ablation}. Removing the Accumulation CLSTM (\emph{w/o acc.}) and Forecasting LSTM (\emph{w/o autoreg.}) leads to a degradation in performance, confirming their effectiveness in modeling temporal dependencies. For the latter, we replace the autoregressive decoder with a single-step multi-frame prediction along the channel dimension. Additionally, we evaluate the impact of reverse motion flow (velocity) by excluding it from the input. Performance decreases in its absence, suggesting that the architecture may be suboptimal at estimating velocity from occupancy alone.

\begin{table}
  \caption{\textbf{WOMD:} Ablation Study for architecture and input choices (validation set)}
  \label{tab:womd_ablation}
  \centering
  \small
  \begin{tabular}{lccccccc}
    \toprule
    {Model} & \multicolumn{2}{c}{Observed} & \multicolumn{2}{c}{Occluded} & & \multicolumn{2}{c}{Flow-Grounded} \\
    \cmidrule(r){2-3} \cmidrule(r){4-5} \cmidrule(r){7-8}
    & AUC & Soft IoU & AUC & Soft IoU & Flow EPE & AUC  & Soft IoU  \\
    \midrule
    Baseline            & 0.7703 & 0.4879 & 0.1391 & 0.0416 & 2.9627 & 0.7893 & 0.5982 \\
    Submission          & 0.7899 & 0.5169 & 0.1429 & 0.0413 & 2.6007 & 0.8008 & 0.6118 \\
    w/o input flow      & 0.7578 & 0.4722 & 0.1366 & 0.0405 & 3.1504 & 0.7821 & 0.5865 \\
    w/o accumulation    & 0.7448 & 0.4526 & 0.1116 & 0.0314 & 3.3193 & 0.7746 & 0.5747 \\
    w/o autoregression  & 0.7469 & 0.4599 & 0.1278 & 0.0366 & 3.4480 & 0.7720 & 0.5825 \\
    \bottomrule
  \end{tabular}
\end{table}

\textbf{AV2:} As AV2 is rasterized in an ego-centric reference frame, we evaluate the benefit of incorporating rasterized IMU data. This input is critical in practical settings, where agent trajectories are conditioned on ego-motion. Without explicit IMU input, the network must infer ego dynamics from static features, a more difficult task. Results show that performance improves with IMU input, indicating that the model leverages this information (see Tab.~\ref{tab:argoverse_ablation} and Fig.~\ref{fig:curve_plot_av2}).

\section{Conclusion}
\label{sec:Conclusion}

We propose a fully convolutional sequence-to-sequence LSTM architecture for occupancy flow forecasting in autonomous driving. Our method achieves competitive performance on the Waymo Open Motion Dataset while maintaining a lightweight design optimal for convolution-specialized Neural Processing Units (NPUs). It avoids reliance on vectorized inputs, making it suitable for end-to-end integration with frameworks using surround-view camera encoders (e.g., SimpleBEV). Autoregressive decoding enables online, variable-length forecasting horizons, and we provide evidence that the model benefits from longer input sequences without incurring additional inference overhead. Experiments on the Argoverse 2 dataset validate the model’s generalizability to ego-centric coordinate systems and its ability to leverage rasterized IMU inputs.

Despite its advantages, the model has inherent limitations. Its spatial receptive field is constrained by the convolutional kernel size, and temporal reasoning is bounded by the memory capacity of the LSTM. Additionally, autoregressive decoding requires sequential inference and restricts outputs to fixed time intervals, which limits temporal sampling of predictions. This design also requires calculating the occupancy and flow for every grid cell, unlike more efficient methods that predict implicit occupancy grids. Due to limitations in input sequence length in the dataset, we could not evaluate performance on very long temporal horizons.

{
\small
\bibliographystyle{plain}  
\bibliography{main}
}


\appendix

\section{Additional Qualitative Results}
\label{appendix:additional_qualitative_results}

\begin{figure}
    \label{figure:images_1}
    \centering
    \begin{minipage}{0.24\textwidth}
        \centering
        \includegraphics[width=\textwidth]{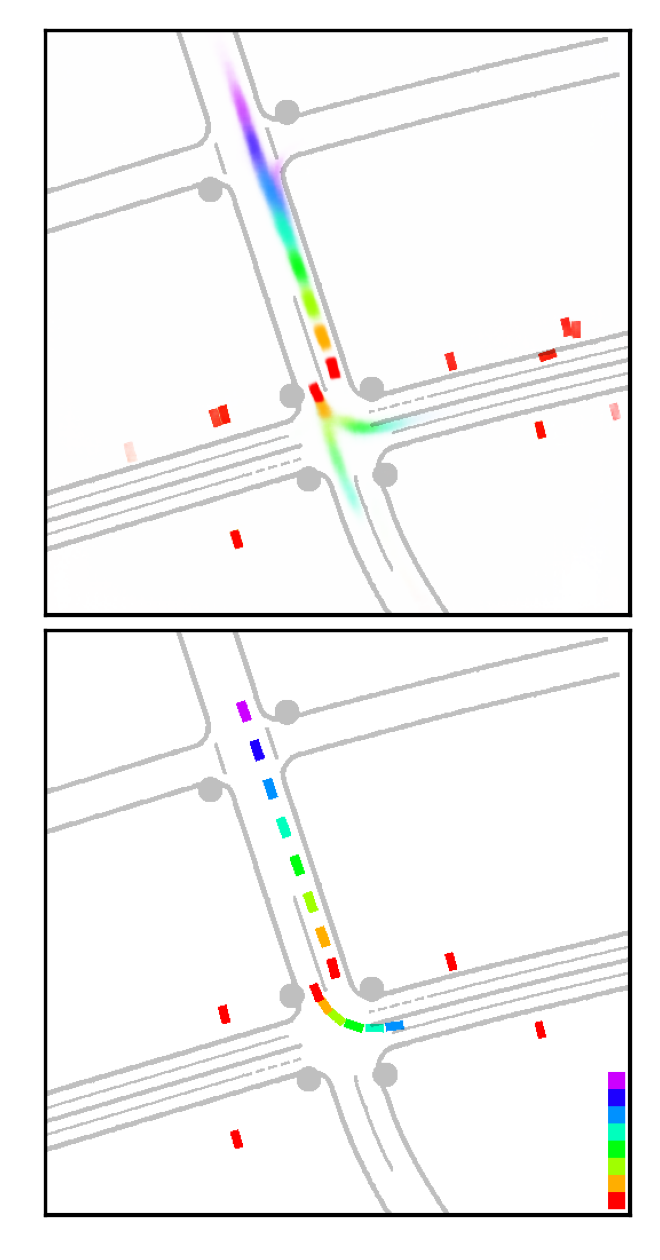}
    \end{minipage} 
    \begin{minipage}{0.24\textwidth}
        \centering
        \includegraphics[width=\textwidth]{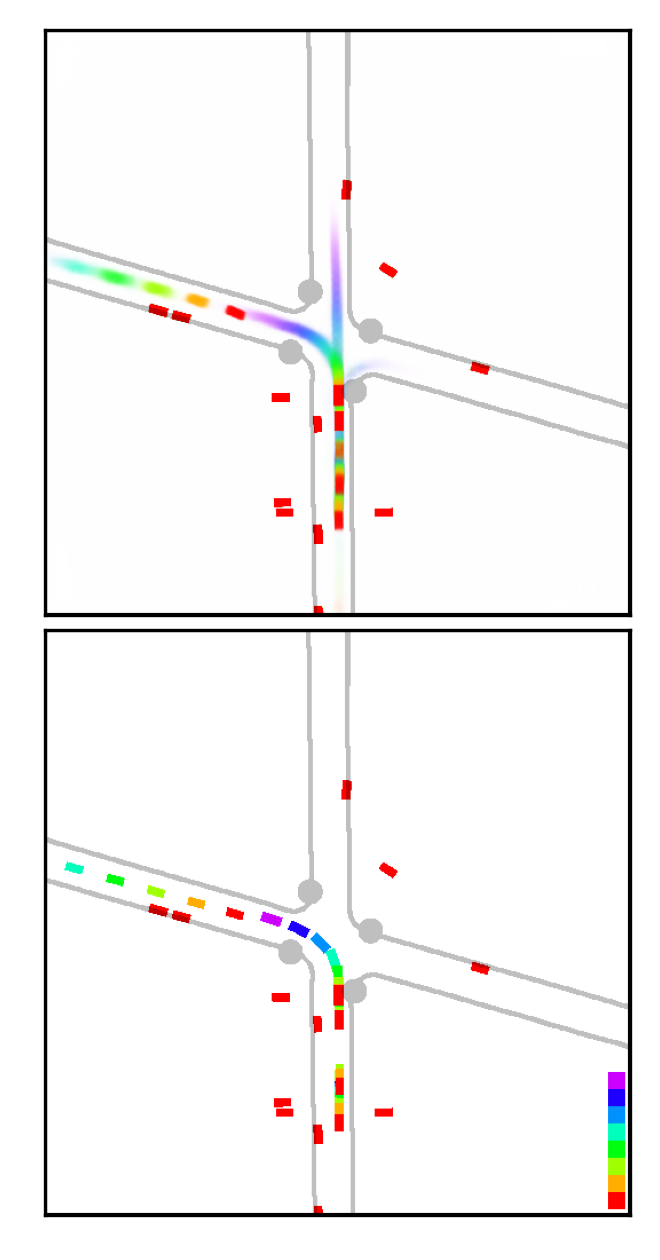}
    \end{minipage}
    \begin{minipage}{0.24\textwidth}
        \centering
        \includegraphics[width=\textwidth]{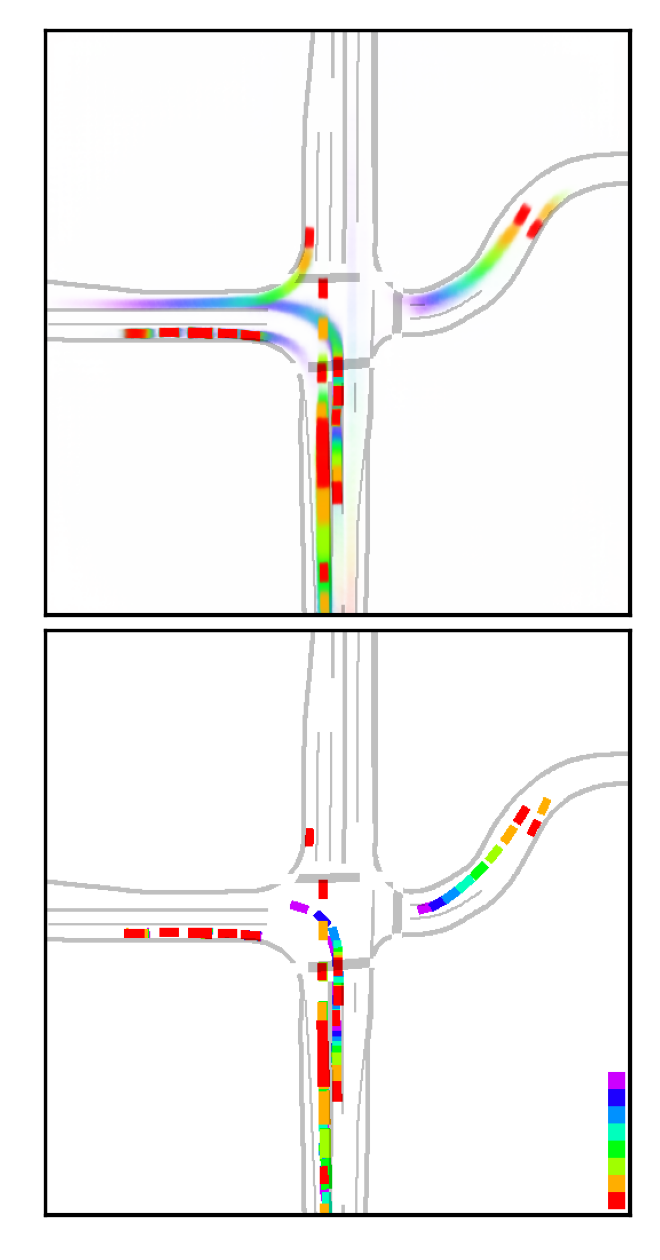}
    \end{minipage}
    \begin{minipage}{0.24\textwidth}
        \centering
        \includegraphics[width=\textwidth]{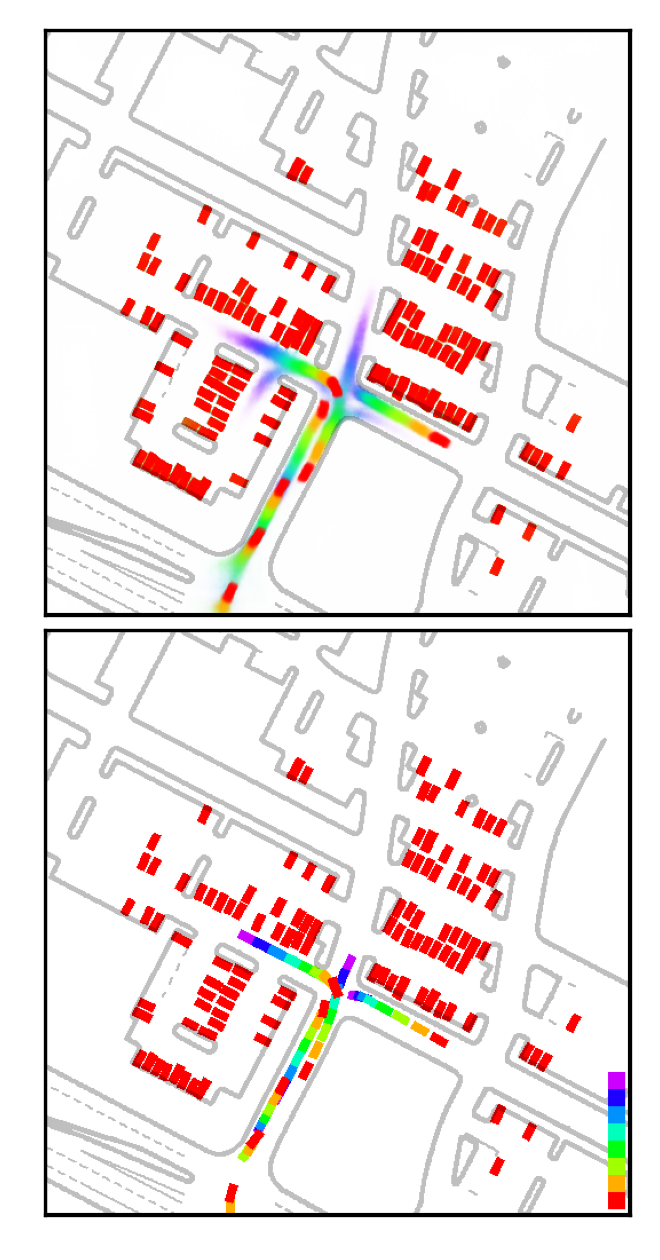}
    \end{minipage}

    \centering
    \begin{minipage}{0.24\textwidth}
        \centering
        \includegraphics[width=\textwidth]{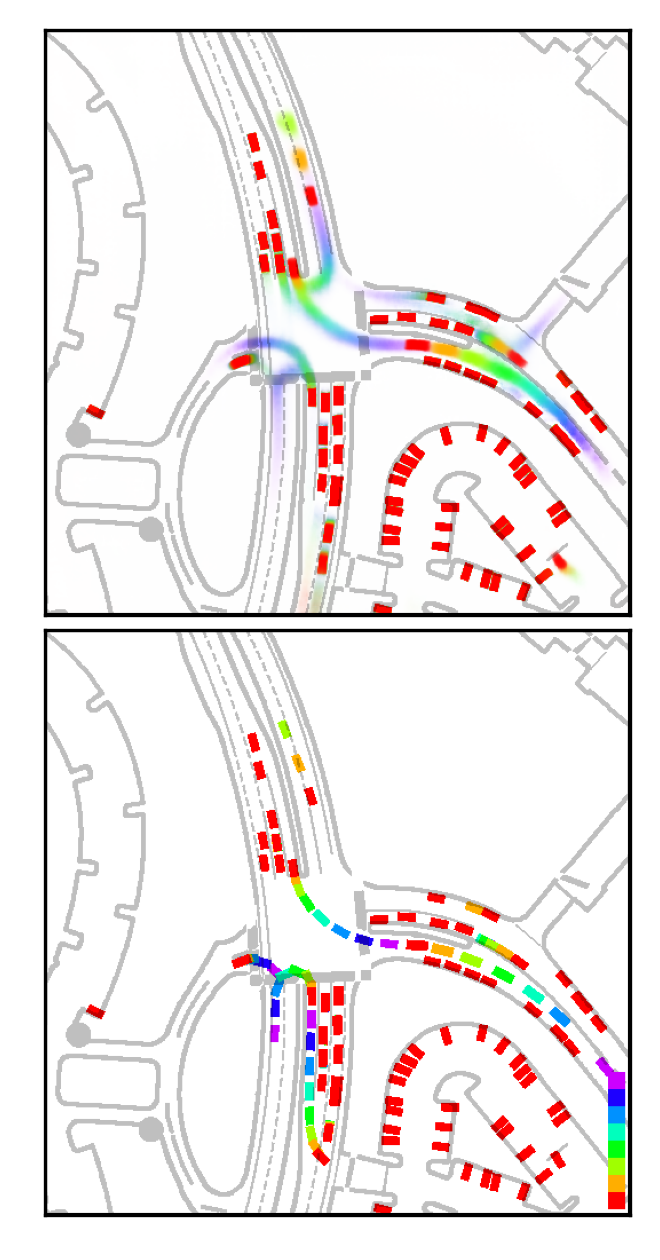}
    \end{minipage} 
    \begin{minipage}{0.24\textwidth}
        \centering
        \includegraphics[width=\textwidth]{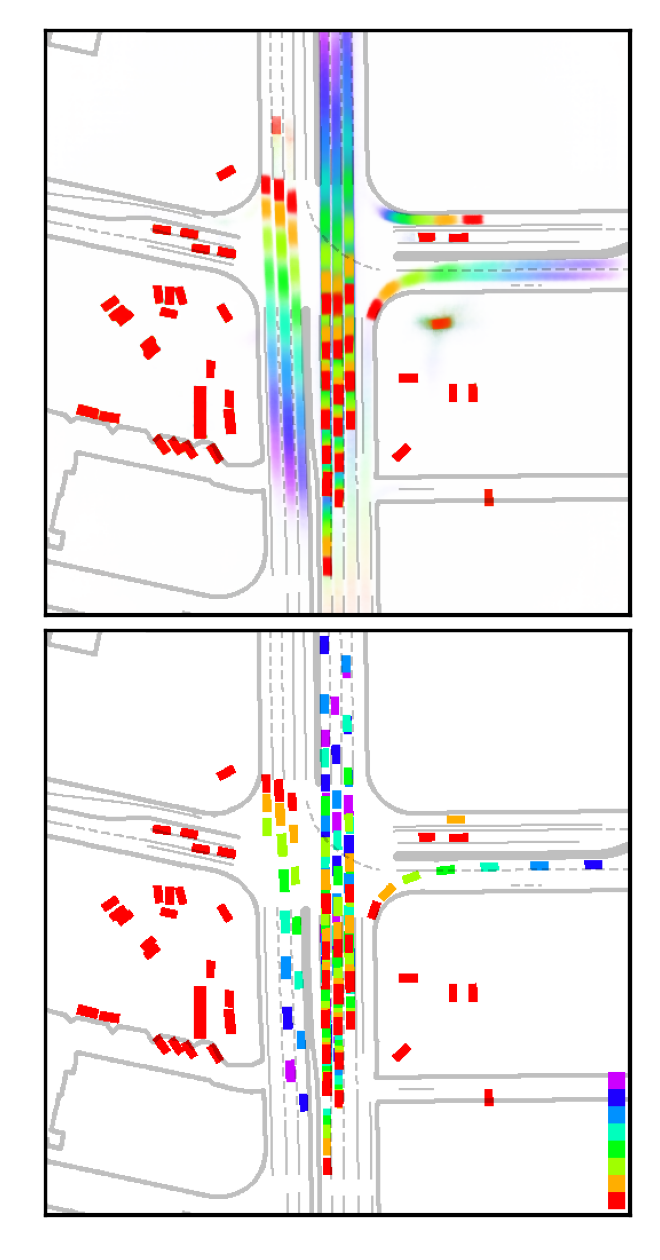}
    \end{minipage}
    \begin{minipage}{0.24\textwidth}
        \centering
        \includegraphics[width=\textwidth]{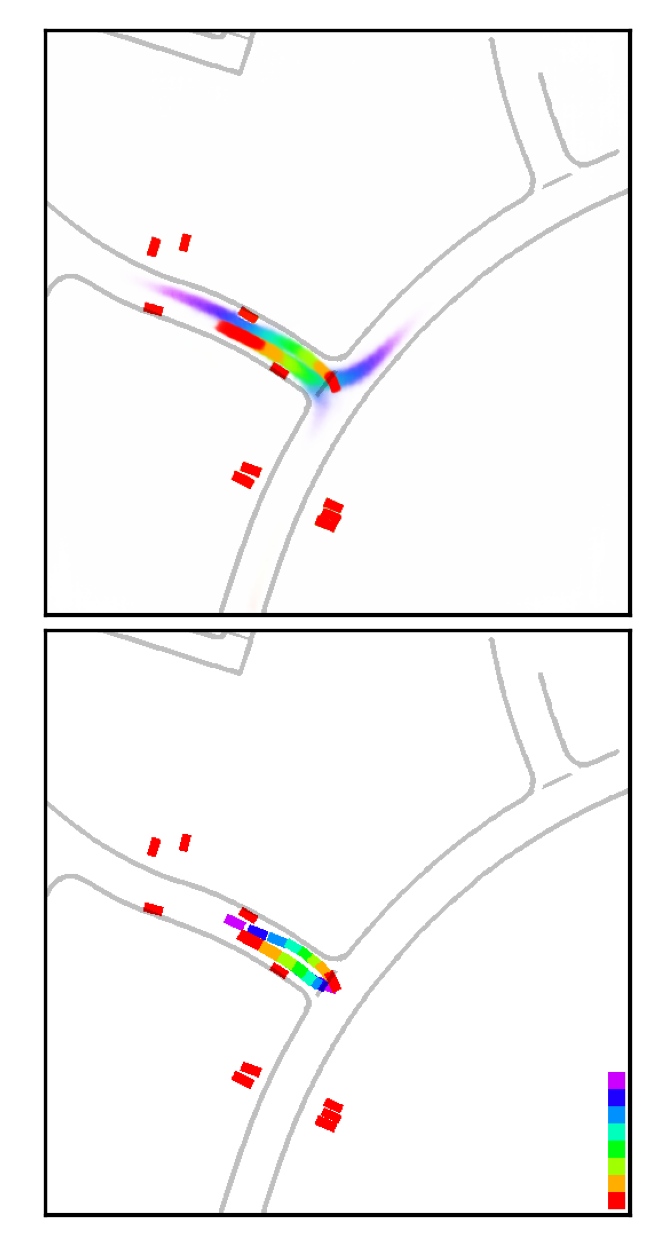}
    \end{minipage}
    \begin{minipage}{0.24\textwidth}
        \centering
        \includegraphics[width=\textwidth]{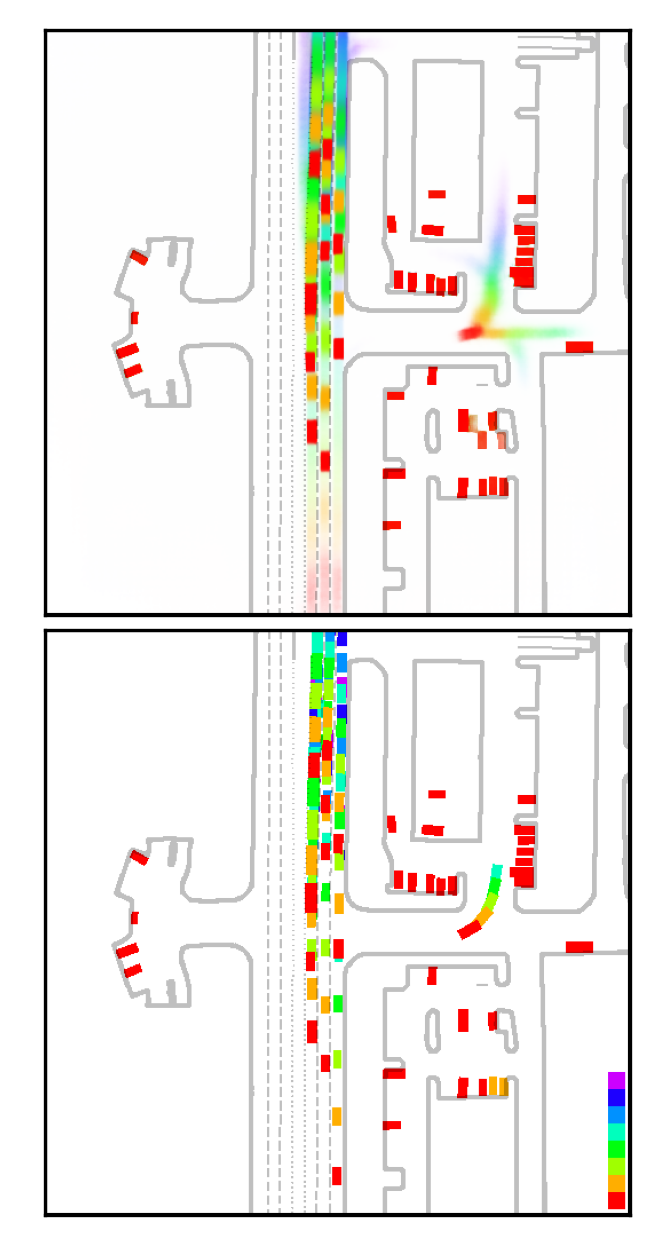}
    \end{minipage}

        \centering
    \begin{minipage}{0.24\textwidth}
        \centering
        \includegraphics[width=\textwidth]{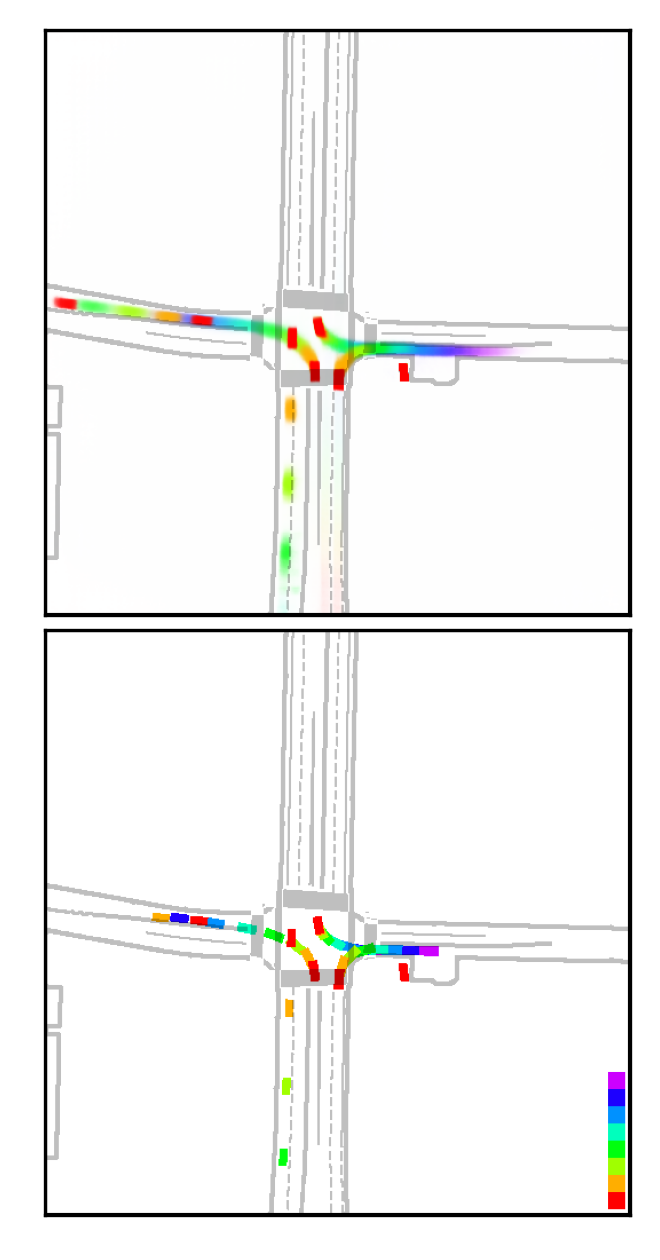}
    \end{minipage} 
    \begin{minipage}{0.24\textwidth}
        \centering
        \includegraphics[width=\textwidth]{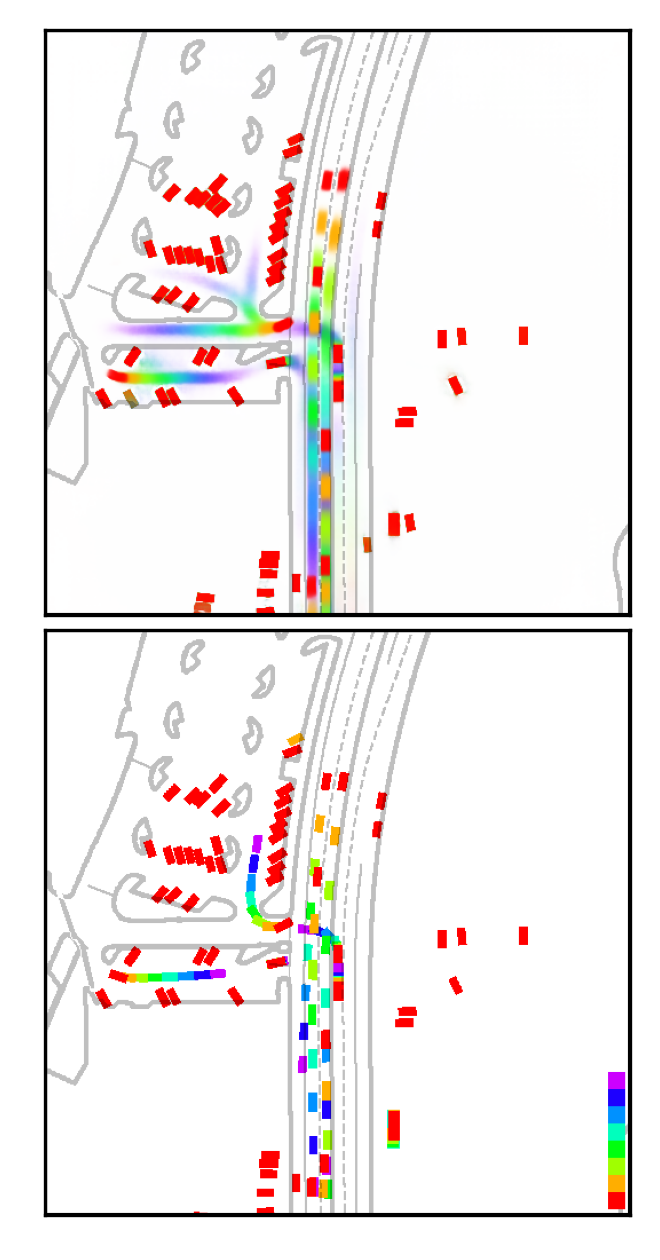}
    \end{minipage}
    \begin{minipage}{0.24\textwidth}
        \centering
        \includegraphics[width=\textwidth]{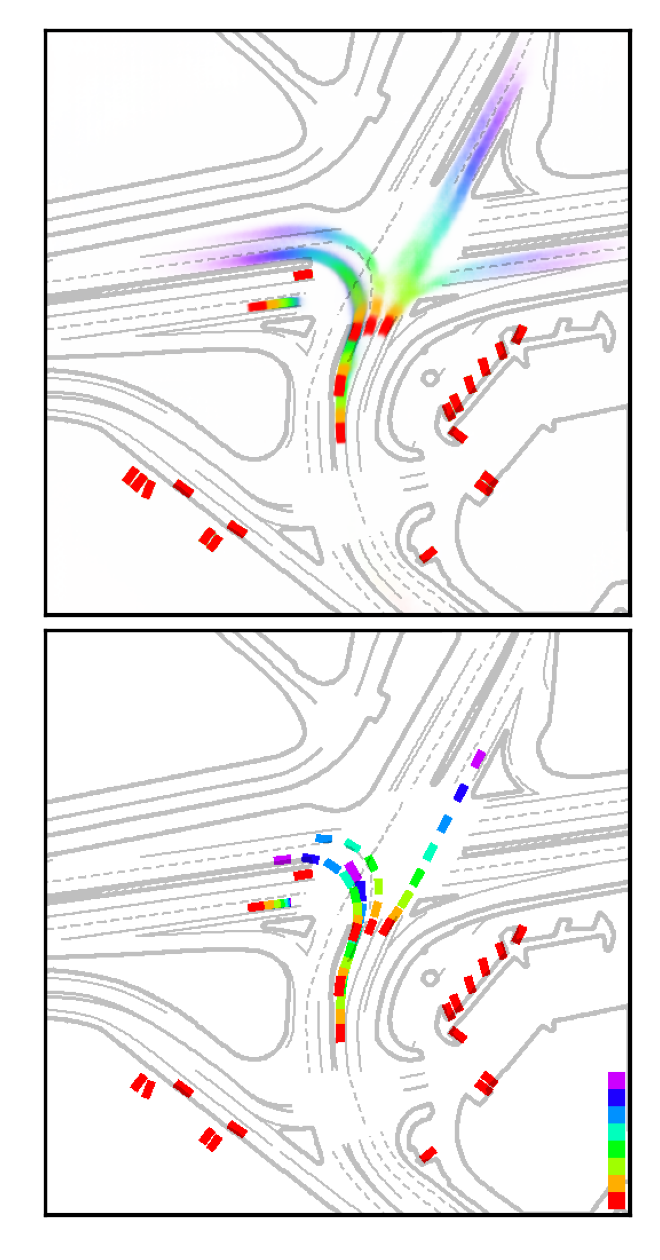}
    \end{minipage}
    \begin{minipage}{0.24\textwidth}
        \centering
        \includegraphics[width=\textwidth]{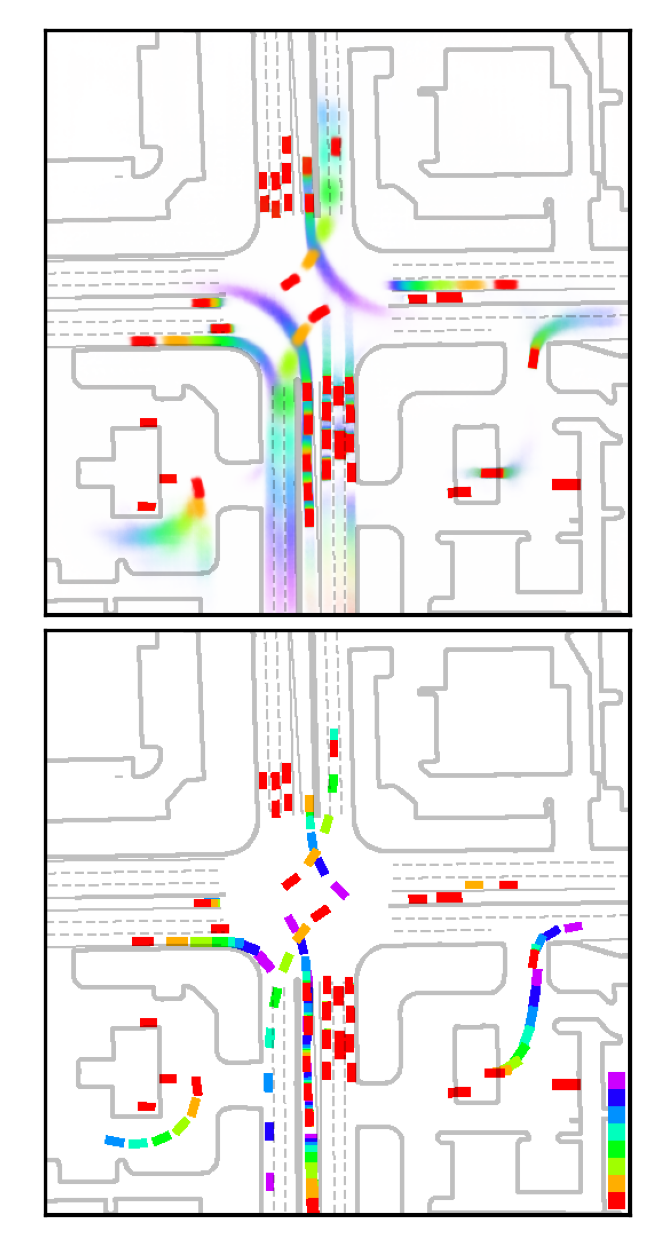}
    \end{minipage}
    
    \caption{\textbf{Qualitative results on WOMD validation set}. Each subplot displays the testing result of (1) color coded future occupancy prediction, (2) color coded future occupancy target. The results are the outputs of our state-of-the-art model using \(H, W = 512\) input rasters. Color coding denotes timesteps \( t \in [1, T_f] \) with \( red = 1 \).}
\end{figure}
\begin{figure}
    \label{figure:images_2}
    \centering
    \begin{minipage}{0.24\textwidth}
        \centering
        \includegraphics[width=\textwidth]{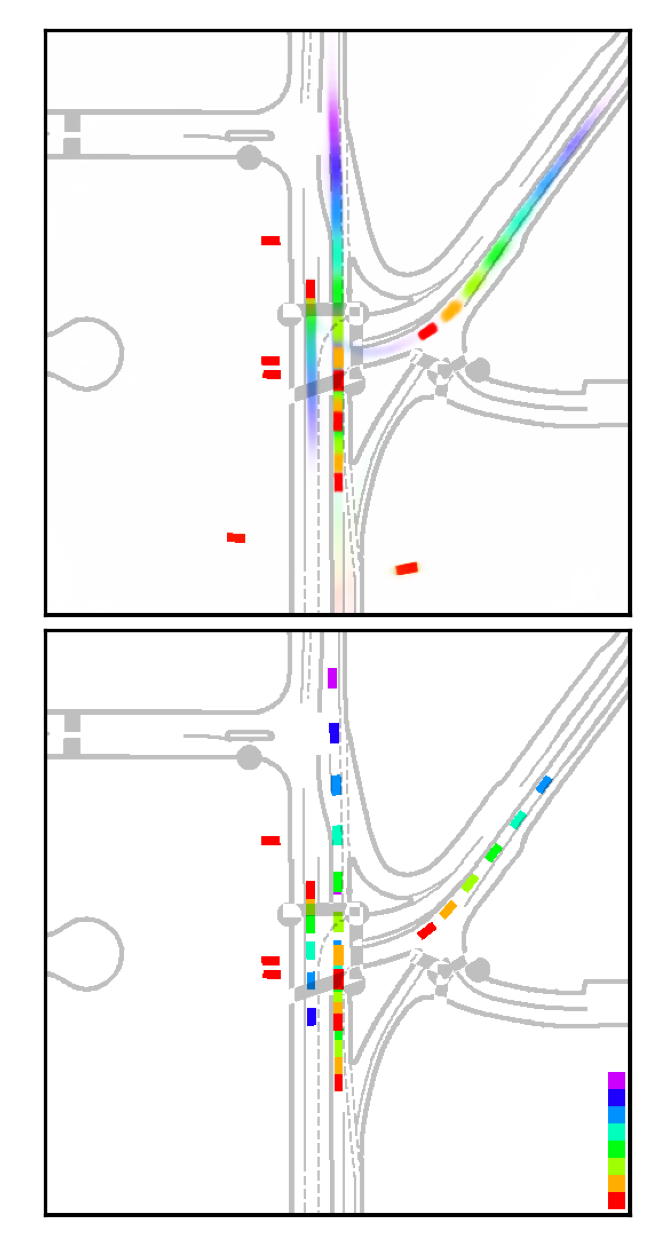}
    \end{minipage} 
    \begin{minipage}{0.24\textwidth}
        \centering
        \includegraphics[width=\textwidth]{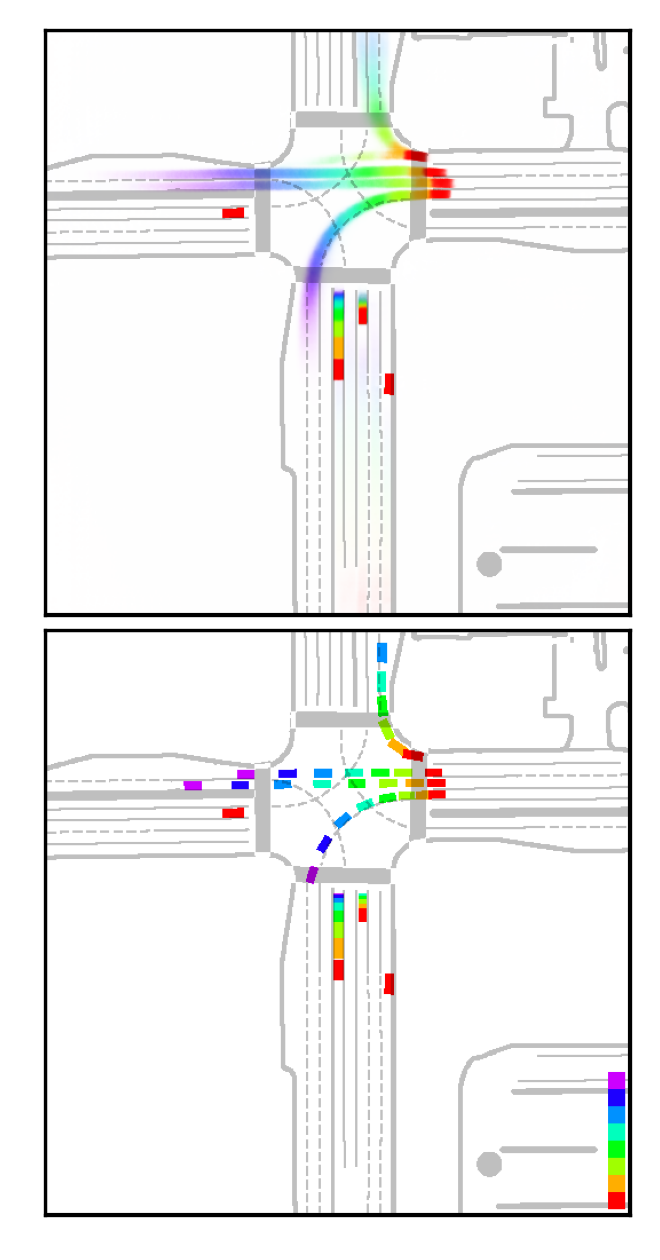}
    \end{minipage}
    \begin{minipage}{0.24\textwidth}
        \centering
        \includegraphics[width=\textwidth]{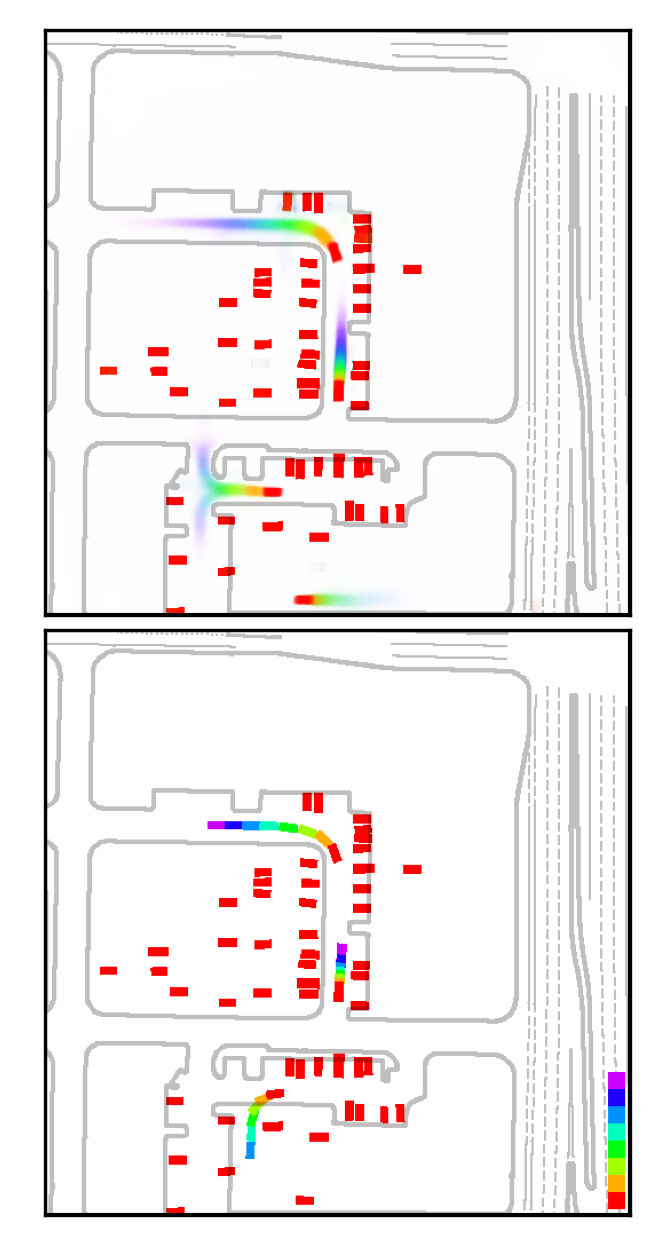}
    \end{minipage}
    \begin{minipage}{0.24\textwidth}
        \centering
        \includegraphics[width=\textwidth]{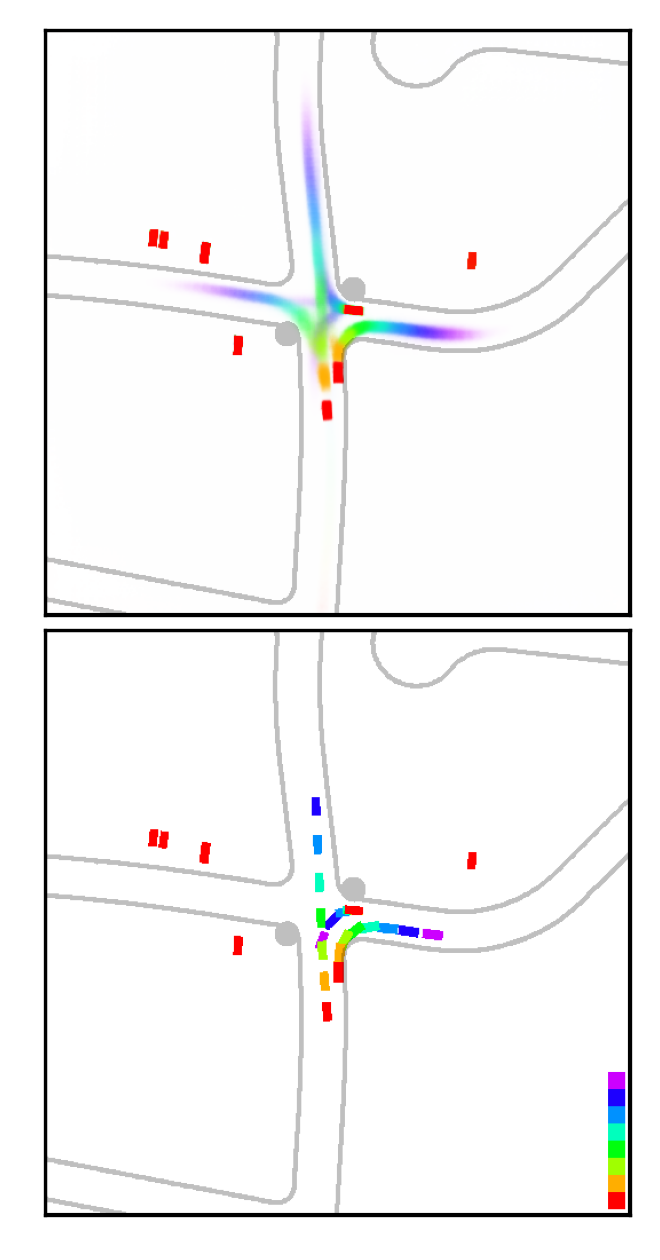}
    \end{minipage}

    \centering
    \begin{minipage}{0.24\textwidth}
        \centering
        \includegraphics[width=\textwidth]{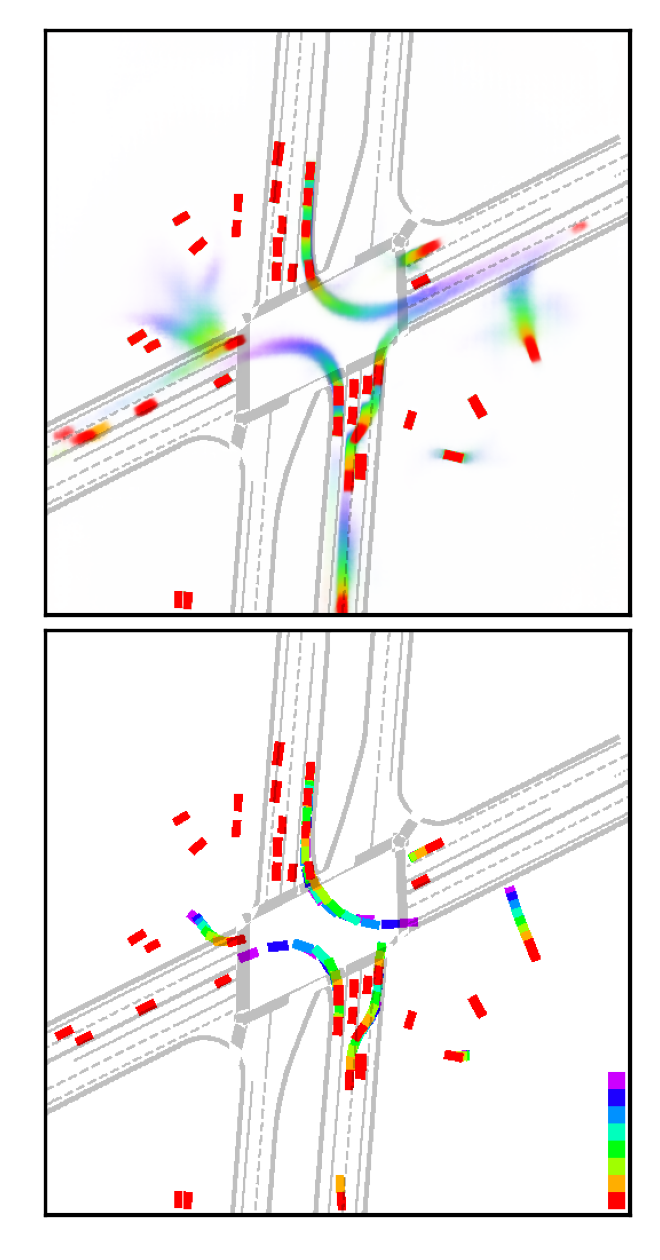}
    \end{minipage} 
    \begin{minipage}{0.24\textwidth}
        \centering
        \includegraphics[width=\textwidth]{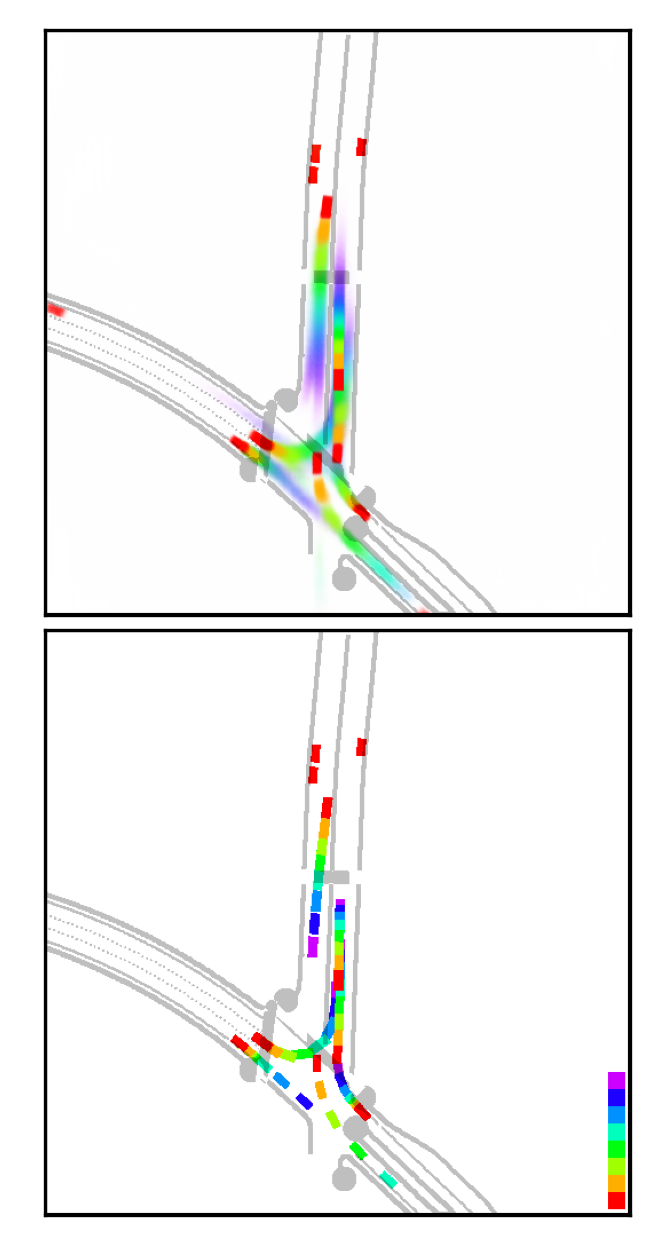}
    \end{minipage}
    \begin{minipage}{0.24\textwidth}
        \centering
        \includegraphics[width=\textwidth]{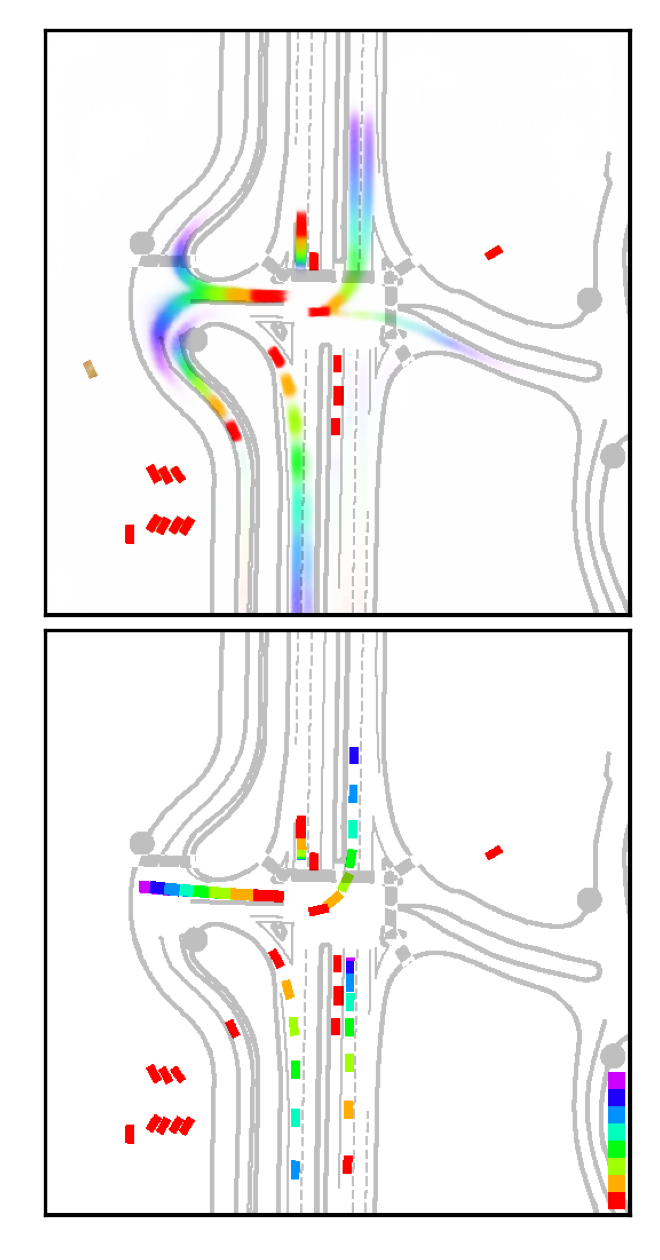}
    \end{minipage}
    \begin{minipage}{0.24\textwidth}
        \centering
        \includegraphics[width=\textwidth]{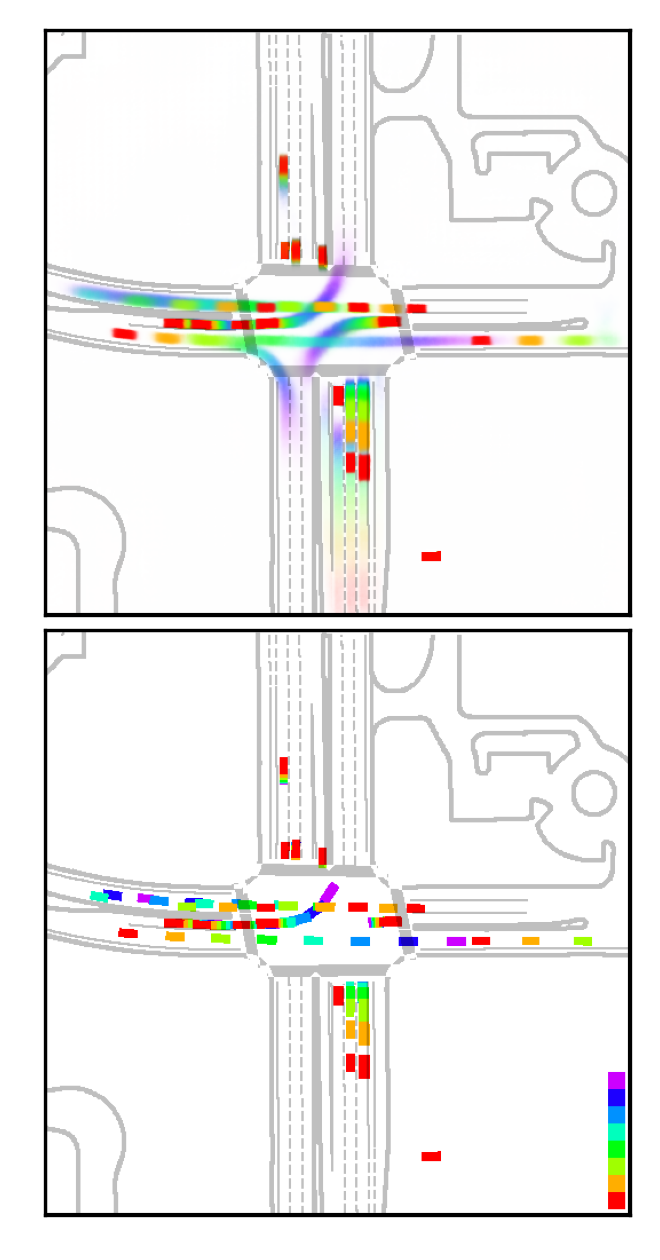}
    \end{minipage}

        \centering
    \begin{minipage}{0.24\textwidth}
        \centering
        \includegraphics[width=\textwidth]{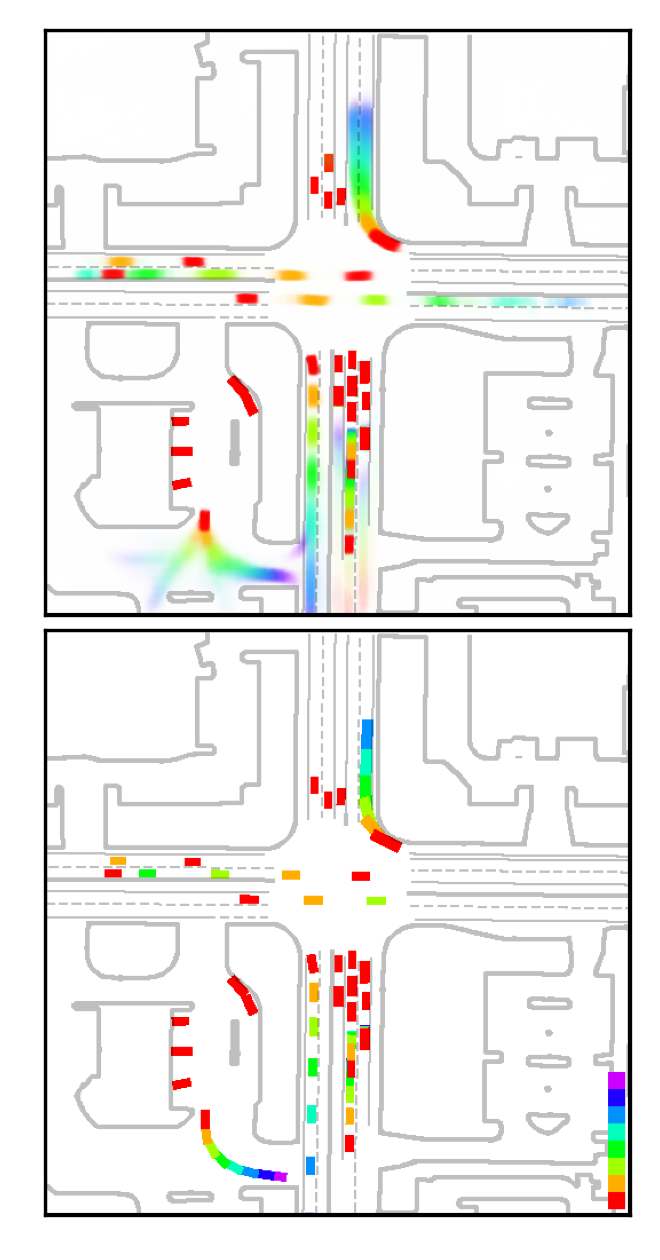}
    \end{minipage} 
    \begin{minipage}{0.24\textwidth}
        \centering
        \includegraphics[width=\textwidth]{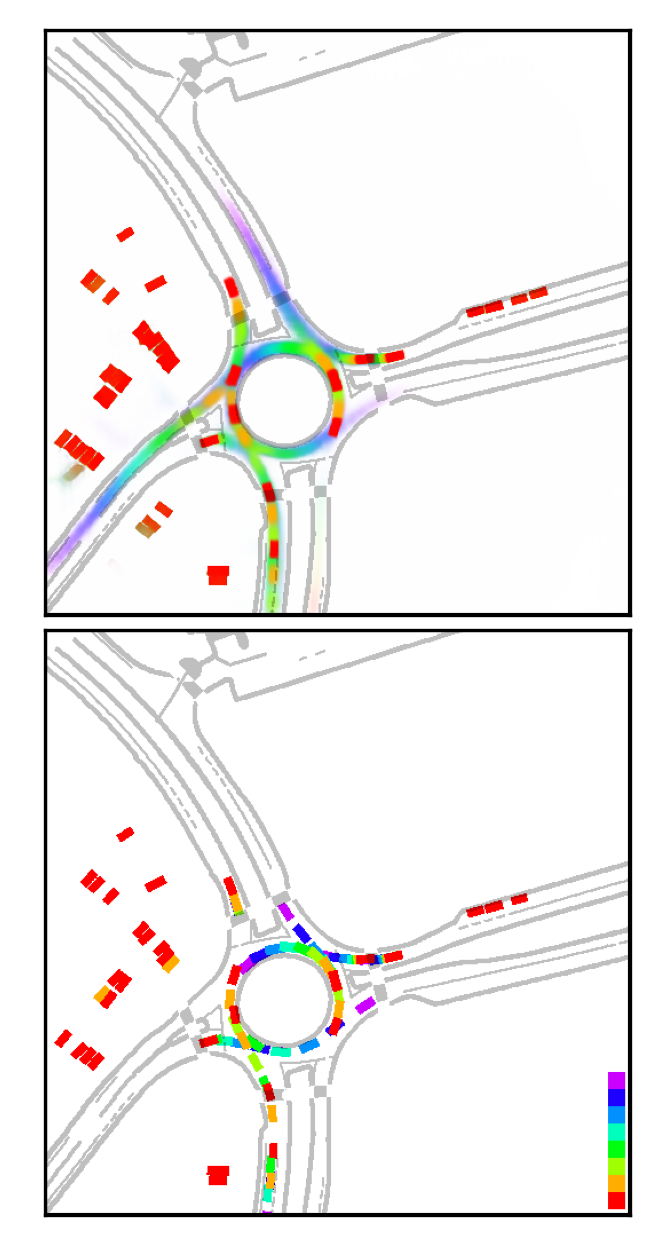}
    \end{minipage}
    \begin{minipage}{0.24\textwidth}
        \centering
        \includegraphics[width=\textwidth]{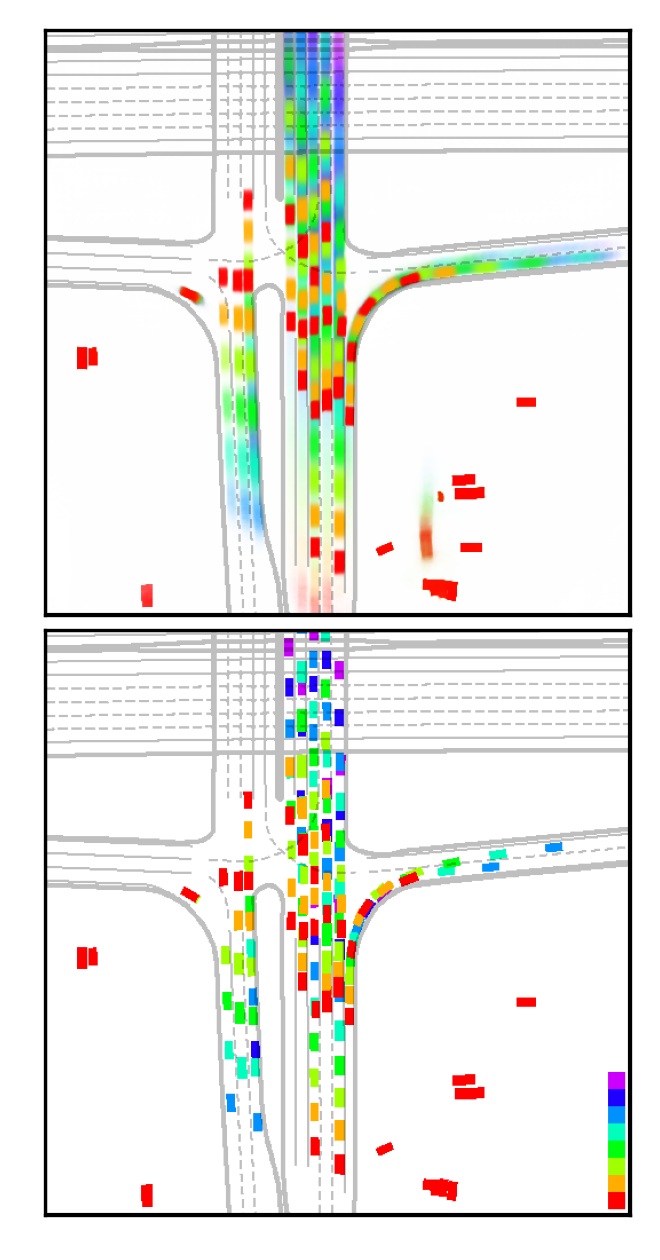}
    \end{minipage}
    \begin{minipage}{0.24\textwidth}
        \centering
        \includegraphics[width=\textwidth]{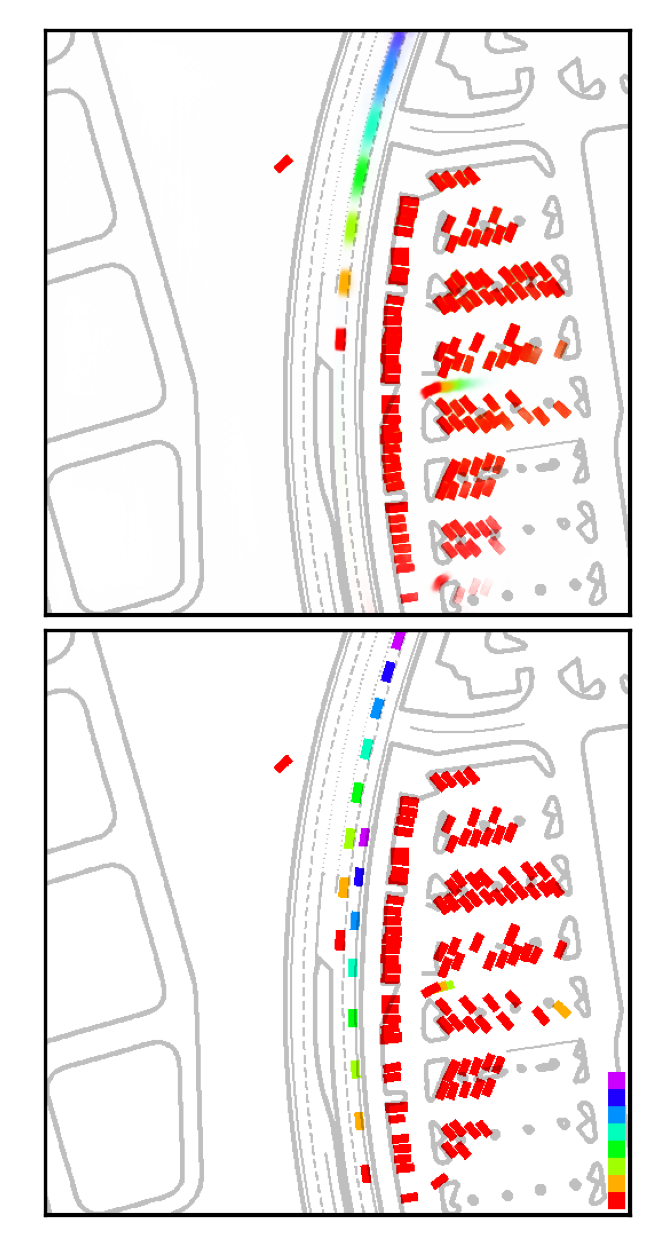}
    \end{minipage}
    
    \caption{\textbf{Qualitative results on WOMD validation set}. Each subplot displays the testing result of (1) color coded future occupancy prediction, (2) color coded future occupancy target. The results are the outputs of our state-of-the-art model using \(H, W = 512\) input rasters. Color coding denotes timesteps \( t \in [1, T_f] \) with \( red = 1 \).}
\end{figure}


\ifpreprint
\fi

\end{document}